 \pdfoutput=1
\documentclass[preprint,12pt,authoryear]{elsarticle}

\usepackage{amssymb}
\usepackage[utf8]{inputenc} % allow utf-8 input
\usepackage[T1]{fontenc}    % use 8-bit T1 fonts
\usepackage{hyperref}       % hyperlinks
\usepackage{url}            % simple URL typesetting
\usepackage{booktabs}       % professional-quality tables
\usepackage{amsfonts}       % blackboard math symbols
\usepackage{nicefrac}       % compact symbols for 1/2, etc.
\usepackage{microtype}      % microtypography
\newcommand{\INDSTATE}[1][1]{\STATE\hspace{#1\algorithmicindent}}
\usepackage{amsmath}  % AMS mathematical facilities for LaTeX
\usepackage{amsfonts}
\usepackage{algorithm}
\usepackage{algorithmic}
\usepackage{wrapfig}
\usepackage{graphicx}
\usepackage{subcaption}
\usepackage{float} % Improved interface for floating objects
\usepackage{graphicx}
\usepackage{caption}
\usepackage{dirtytalk}
\usepackage[threshold=0]{csquotes}
\setcitestyle{aysep{ }}
\usepackage{xcolor}

\journal{arXiv}

\begin{document}

\begin{frontmatter}


% \author{Name\corref{cor1}\fnref{label2}}
% \address{Address\fnref{label3}}

\title{Learning Feature Relevance Through Step Size Adaptation in Temporal-Difference Learning}

\author[label1,label2]{Alex Kearney}
\author[label1,label2]{Vivek Veeriah}
\author[label2]{Jaden Travnik}
\author[label2,label3]{Patrick M. Pilarski}
\author[label2,label3]{Richard S. Sutton}
\address[label1]{Borealis AI, Edmonton, Alberta, Canada}
\address[label2]{University of Alberta, Edmonton, Alberta, Canada}
% \address[label3]{DeepMind, Edmonton, Alberta, Canada}

% maybe roll into everything the work
\begin{abstract}

There is a long history of using meta learning as representation learning, specifically for determining the relevance of inputs. In this paper, we examine an instance of meta-learning in which feature relevance is learned by adapting step size parameters of stochastic gradient descent---building on a variety of prior work in stochastic approximation, machine learning, and artificial neural networks. In particular, we focus on stochastic meta-descent introduced in the Incremental Delta-Bar-Delta (IDBD) algorithm for setting individual step sizes for each feature of a linear function approximator. Using IDBD, a feature with large or small step sizes will have a large or small impact on generalization from training examples. As a main contribution of this work, we extend IDBD to temporal-difference (TD) learning---a form of learning which is effective in sequential, non i.i.d. problems. We derive a variety of IDBD generalizations for TD learning, demonstrating that they are able to distinguish which features are relevant and which are not. We demonstrate that TD IDBD is effective at learning feature relevance in both an idealized gridworld and a real-world robotic prediction task. 

% large or small sucks
\end{abstract}

% todo: ask rich about capstone sentences


\end{frontmatter}
% \linenumbers

\section{Representation Learning Through Feature Relevance}
\label{Introduction}
% is it clear what we're talking about?
% are terms and concepts properly introduced?
% are you making jumps between thoughts?

% 1) representations are important & we would prefer to automate finding them

% - the performance of a system is highly dependent on the features a system uses to represent it's environment. 

% - the difference between good and bad features can be the difference between a system which is successful and one which is unable to learn

% - features are so important that substantial effort is put into choosing good features

The performance of Machine Learning (ML) methods depends on their representation of the environment. A representation is composed by a collection of features which describe aspects of the current state or example which is observed by the system. The choice of representation can be the difference between a successful application and one which is unable to learn. For this reason, representations are often designed by engineers who construct features which best describe the task at hand. 

%To construct appropriate features, the designer must determine which aspects of the environment are relevant to the particular problem from a set of available inputs.
%  Aside: rich was thinking of going into feature learning, and rep'n learning, but I fear that in doing so we're going to add more confusion by muddling what we're talking about

% 2) wouldn't it be just better if we automated it?

% - cite bengio's rep'n learning thing

% - Hand-constructing features is limited and does not scale 

% - For this reason, we would like to be able to find features automatically

% - There have been trends within the community from  hand-construction to automatic discovery of rep'ns and features; this has in turn led to substantial progress 

Effectiveness of hand-constructed of features is limited, as designing a representation requires substantial knowledge of both the environment and the problem being solved. The features which are appropriate for a given task are not necessarily transferable to different environments and problems. For each new problem and environment the engineer must re-design the representation. For this reason, hand-constructed features and engineer expertise is not a scalable approach to representation construction.  Due to these limitations, it is desirable for systems to independently and automatically construct the features a system uses to learn. 

% 3) The process of automating representation generation can itself be learned through a second-order learning process. 

% - we can learn our automation of feature discovery through a second-order learning process

% - meta learning and learning to learn: what counts as it

% - constructing features and learning relevance as the two main ways of modifying a representation

The process of automatically deciding how to represent the environment can itself be learned through a second-order learning process---sometimes described as representation learning, or more broadly as both \textit{learning to learn} and \textit{meta-learning}. Meta-learning methods learn to modify the inputs or the parameters of the underlying machine learning method. We can use meta-learning to perform representation learning by learning to constructing new features, or learning to shaping an existing representations by weighting given features. In this paper, we focus on systems which perform meta-learning in order to shape a given representation, i.e. by identifying relevant features. 

% 4) what is feature relevance?

%  - the following may muddle the argument 

%  - establish feature relevance

%  - how can we learn feature relevance

The simplest case of representation learning is learning the relevance of given features. Whether features are constructed by hand or learned, the features a system uses will not be equally relevant to the task at hand. There are some features which will be more relevant and which we wish to generalize over more than others.  By learning feature relevance, a system can weight the influence of input features on the model being learnt and generalization from training examples. In this sense, learning the relevance of features is a form of representation learning which is prior to feature construction: before creating systems which are able to construct their own features from inputs, a system can modify its existing representation by weighting given features based on their relevance.
%  5) argument from animal learning Rich's unpublished paper, yael niv, keston, DBD, yellow report, IDBD
%  not just something that's in ML; there is a basis for this in natural learning
%  work outside of machine learning has studied how humans are able to determine which stimuli are important
%  work outside ofmachine learning has studied how humans are able generalize from a few examples by learning the relevance of inputs
The problem of identifying relevant features through meta-learning has roots in both animal and human learning. Humans and animals are known to learn to discern which aspects of the environment are relevant to the task at hand. Work in neuroscience has assessed how humans perform representation learning by identifying relevant stimuli \citep{wilson_inferring_2012}; in cognitive science studies have examined how children are able to generalize from just a few examples by forming appropriate biases\citep{colunga_lexicon_2005}. Animals learn over which features to generalize their learning to new examples--they learn the saliency of the signals. In doing so, humans and animals are performing representation learning  by identifying the relevance of stimuli. It is natural that a machine learning system perform representing learning by learning the relevance of inputs.
% todo: cite the saliency thing from rich.

% 5) The simplest way to do this is in a linear case with step size adaptation
% - one feature for each weight, one step size for each feature
% - how does this work?
% - play out an example
% - one way of learning step sizes in sucha fashion is by doing 

One method of assigning feature relevance is through adapting a vector of many step sizes. Step sizes scale updates made to a learned model; by assigning step sizes on a per-feature basis we are able to give large step sizes to relevant features and small step sizes to the irrelevant features: scaling weight updates based on the quality of input features \citep{sutton_adapting_1992}. The simplest case of step sizes as feature relevance are methods which use a linear function approximation: methods where each weight has its own specific step size. In this paper, we focus on methods which use linear function-approximation. In  the linear mapping case there is a single step size per feature and feature-relevance directly corresponds with step sizes. One of the main methods of learning feature relevance by learning setting step size values is through Stochastic Meta-descent (SMD): a form of gradient descent which takes the gradient of a gradient step. 

%Feature relevance is the simplest case of representation learning. When we learn the relevance of given input features, we are learning how to weight available inputs. This acts as a form of representation learning by enabling modulation of learning based on the quality of given features. Learning a representation through feature relevance is the most basic form of representation learning which can be performed. Weighting provided features based on relevance is a form of representation learning which is prior to decisions about how to construct features: before you create features you must learn their relevance. Even if features are independently constructed, you must still learn the relevance.

\section{Stochastic Meta-descent For Meta Learning And Feature Relevance}

% 1) this is one of the main things that's being done. A key approach to approach to setting the step size values in SMD
% check the claim. Is there _really no prior SMD_?
% we need to develop sufficient intuition

% what is stochastic meta descent
% how does it work?
% second-order methods

% 2) history of SMD where does it come from?
% where was it first introduced
% how has it been used since?

% 3) Biolo

SMD was first introduced for online learning in the linear case as Incremental Delta-Bar-Delta (IDBD)\citep{sutton_adapting_1992} which was later extended to non-linear mappings \citep{schraudolph_local_1999}. Most recently, SMD has been used for temporal predictions and sequential problems, such as MAML \citep{finn_model-agnostic_2017}: an offline method for updating the step sizes in a model-agnostic fashion. Meta-gradient RL uses SMD to adapt the return of a reinforcement learning problem \citep{xu_meta-gradient_2018}, and Cross-prop uses SMD to learn weightings of inputs to learn representations which generalize across tasks.  Outside of machine learning, IDBD has been extended to biologically plausible version for modeling neural metaplasticity \citep{schweighofer_model_1998}.  

%  3) other things which are linear and performing step size adaptation and feature relevance
Many generalizations of IDBD have been used to learn step sizes online for a variety of underlying learning methods. AutoStep \citep{mahmood_representation_2013} is an extension of IDBD which introduces a normalization to the update of IDBD's meta-weights to improve stability. SID, NOSID, \citeauthor{dabney_adaptive_2014} AutoStep \citep{dabney_adaptive_2014}, and our initial presentation of TIDBD \citep{kearney_every_2017} are generalizations of IDBD for TD methods. We are particularly interested in these meta-learning methods for step size adaptation to learn the relevance of features. 

\section*{IDBD}

% this introduction needs to be re-worked to fit with theme
% 

Incremental Delta-Bar-Delta(IDBD) \citep{sutton_adapting_1992} is a meta-learning algorithm which learns a bias through experience by maintaining a vector of learned step sizes. The intuition behind IDBD is that features which are correlated with our prediction problem should be given large step sizes, while features which are irrelevant to our prediction problem should be given smaller step sizes.

IDBD estimates a target value $\hat{y}$ by learning a weight vector $w$, such that the dot-product of the observations and the weight vector produce an estimate $y = w^\top x$. The weights are updated by moving in the direction of the error $\delta = \hat{y} - y$, where $\hat{y}$ is the observed target. The update is weighted by a small fraction $\alpha > 0$, where $\alpha$ is a vector of step sizes which are learnt and $\alpha_i > 0$. 

IDBD adapts a vector of many step sizes online and incrementally by performing stochastic meta-descent over a vector of meta-weights $\beta$ which are used to specify the step sizes $\alpha$ on a per-feature basis. On each time-step alpha is updated by $\alpha \gets \exp(\beta)$, producing a vector of step sizes: one step size for each feature in observations $x$. By exponentiating the meta-weights $\beta$ to produce a step size $\alpha$, a linear step in the meta-weights $\beta$ produces a geometric step in $\alpha$; in addition, it ensures all step sizes $\alpha$ are positive.  

The meta-weights are updated by $\beta \gets  \beta + \theta \delta x h$. The meta step size $\theta$ weights the amount by which we update our weights in the direction of the error. The prediction error $\delta$ is the difference between our estimate $y$ and the observed target $\hat{y}$.  The additional memory vector $h$ is a decaying trace of recent meta-weight updates. The update of the $\beta$ is proportional to the correlation between the current change and previous changes to $\beta$. 

The intuition behind this update is that if many updates for feature $x_i$ to weight $w_i$ are similar, it would have been a more effective use of data to have a larger step size $\alpha_i$ and thus a larger update to $w_i$. Negative correlation suggests that updates to $w_i$ have over-shot and over-corrected---that the step size $\alpha_i$ should be smaller. 

% maybe add a figure of converging to optimal values

\begin{algorithm}[ht]
\caption{IDBD}
\label{IDBD}
\begin{algorithmic}[1]
\STATE Initialize vectors $h$, $\beta$, and $w$ of size $n$ number of features.
\STATE Repeat for each observation $x$ and target $\hat{y}$: 
  \INDSTATE[1] $y \gets w^\top x$
  \INDSTATE[1] $\delta \gets \hat{y} - y$
  \INDSTATE[1] For $i = 1, 2, \cdots, n $: 
  	\INDSTATE[2] $\beta_i \gets \beta_i + \theta \delta x_i h_i$
    \INDSTATE[2] $\alpha_i \gets e^{\beta_i}$
  	\INDSTATE[2] $w_i \gets w_i + \alpha_i \delta x_i$   % If not using traces, \phi(s_t)$
  	\INDSTATE[2] $h_i \gets h_i [1-\alpha_i x_i^2]^+ + \alpha_i\delta x_i$ 
\end{algorithmic}
\label{TIDBD_alg}
\end{algorithm}

While it is presented alongside the underlying gradient-descent learning mechanism, IDBD is truly the update and maintenance of $\beta$, $\alpha$, and $h$ (lines 6,7, and 9, respectively). In this sense, it is a meta-learning method which is distinct and separate from the method which learns resulting model; however, the order of IDBD's updates in relation to the underlying learning updates is important.

A possible criticism of IDBD is that it is only abstracting the problem of setting the step sizes to a higher level: although IDBD learns the step size parameter, we must now specify the meta step size $\theta$ with which we learn our step sizes. This is still an improvement, as tuned IDBD outperforms methods which do not adapt their bias \citep{sutton_adapting_1992}. In addition, extensions of IDBD, including AutoStep \citep{mahmood_tuning-free_2012}, and NOSID, and AUTOSID \citep{dabney_adaptive_2014} have had success in making their methods relatively invariant to the setting of the meta step size $\theta$.  

Another criticism of IDBD is that of stability. If many consecutive weight updates to $w_i$ are in the same direction, then $h_i$ will grow correspondingly large. As $h_i$ grows, $\beta_i$ grows, and $\alpha_i$ increases geometrically in size. This could lead to instability as a single large update to $\beta_i$ could lead to divergence. To prevent this, \cite{sutton_adapting_1992} originally suggests that updates to meta-weights $\beta$ are limited $\pm2$ and that each step size $\alpha_i$ is limited to some maximum value. While this resolves the instability of IDBD, it introduces two thresholds to the algorithm. This adds greater complexity to the practical use of IDBD, as we must pick appropriate thresholds for each problem. while IDBD abstracts the problem of tuning, it does not fully escape it.

%Prediction problems are of importance to many real-world applications. For example, predictions about the environment can inform a robot's decisions and form a key component of their control systems \citep{edwards_machine_2016, sherstan2015collaborative}. Predictions are a way for machines to anticipate events in their world and inform their decision-making processes. 

\section{Temporal Difference Learning}
\label{MRP}
% where is the KNN clustering citation which keston

A limitation of IDBD is that it is derived for supervised learning, preventing its use on other learning problems which update their models with different underlying learning methods. Most existing extensions of IDBD are generalizations to other learning methods. In particular, reinforcement learning policy evaluation methods have received many generalizations: for instance, SID, NOSID, and an AutoStep variation \citep{dabney_adaptive_2014}.

% need a glue sentence

A form of policy evaluation is Temporal-difference learning (TD). TD methods are of note, as they are able to learn directly from experience, and their estimates are learned using \textit{bootstrapping}---TD methods are able to update their estimate $V(s)$ based on their current estimate. Temporal-difference learning methods are driven by the difference between estimates of successive states.

We can think of TD predictions as estimating the value of a state in a Markov Reward Process (MRP). A MRP is described by the tuple \\$<S,p,r,\gamma>$ where $ S $ is the set of all states, $ p(s^{\prime} | s) $ describes the probability of a transition from a state $ s \in S $ to a new state $ s^{\prime} \in S $, $ r(s,s^{\prime}) $ describes the reward observed on a transition from $s$ to $s^{\prime}$, and $0 \le \gamma \le 1$ is a discount factor which determines how future reward is weighted.

\pagebreak

The goal in an MRP is to learn a value function $V(s)$ which estimates the expected return from a given state $v^{*}(s):= \mathbb{E} \{ G_t | S_t = s \}$, where the return is $G_t := \sum^{\infty}_{i=1} \gamma^{i-1} R_{t+i}$, or the discounted sum of all future rewards. Within the context of an MRP, a prediction is an estimation of the value of a state---an accumulation of discounted future reward. For example, the prediction signal could be the position of a robot's gripper, which is later used as an input into a robot's control system.

We describe TD learning with eligibility traces and linear function approximations. The estimate $V(s)$ is learnt by updating a vector of weights $w$. Using some linear function approximation--i.e.: Tile Coding \citep{sutton_reinforcement_1998} or Selective Kanerva Coding \citep{travnik_representing_2017}---we can generate a binary feature vector which represents state $s$. The estimated value of state $s$ is then $V(s) = w^\top\phi(s)$. 

The error we minimize is the TD error $\delta = R + \gamma w^\top \phi(s^{\prime}) - w^\top \phi(s)$.  Note, $R + \gamma w^\top \phi(s^{\prime}) $ is bootstrapped estimate of the true return $G$. We estimate the true return by taking the sum observed reward $R$ and the discounted estimate of the following observed state $V(s^\prime) = \gamma w^\top \phi(s^{\prime}) $. This means that we are learning our estimate of the target $G$ through experience. The discounting factor is $0 \leq \gamma \leq 1$.

The vector $z$ is a decaying trace of recently activated features. At each time step, the trace is decayed by $\gamma \lambda$ and then incremented by the most recently activated features. The parameter $0 \leq \lambda \leq 1$ describes the rate at which we want to decay our traces. This enables current rewards to be attributed to previously visited states. The weights $w$ are then incremented in the direction of the TD error $\delta$, weighted by a small, positive step size $\alpha$, and our eligibility traces $\lambda$.

% todo: better description of elegibility traces. 

\begin{algorithm}[ht]
\caption{TD($\lambda$)}
\label{TD_alg}
\begin{algorithmic}[1]
\STATE Initialize vectors $z \in {0}^{n} $, and both $w \in \mathbb{R}^{n}$; initialize a small scalar $\alpha > 0$; observe state $S$
\STATE Repeat for each observation $s^{\prime}$ and reward $R$: 
  \INDSTATE[1] $\delta \gets R + \gamma w^\top \phi(s^{\prime}) - w^\top \phi(s)$
  \INDSTATE[1] For $i = 1, 2, \cdots, n $: 
  \INDSTATE[2] $z_i \gets z_i  \gamma \lambda + \phi_i(s)$
  \INDSTATE[2] $w_i \gets w_i + \alpha_i \delta z_i$ 
  \INDSTATE[2] $s \gets s^{\prime}$
\end{algorithmic}
\label{TD_alg_semi}
\end{algorithm}

TD prediction methods are effective in life-long continual learning systems---systems where the true return $G_t$ may never be observed, or where the dynamics of they system are too complex to model.  While useful, TD learning has received relatively little interest in step size adaptation methods. Those which almost exclusively adapt, scalar step sizes: they do not adapt bias by performing representation learning. In this paper, we provide a step size based bias adaptation method by generalizing IDBD to TD learning.

% todo: why has this received relatively little interest.
% todo: need a better bridge between the sections

\section{TIDBD: TD Incremental Delta-Bar-Delta}

We now generalize IDBD to TD learning. There exist four other generalizations of IDBD for policy evaluation: Scalar Incremental Delta-Bar-Delta (SID), Normalized Scalar Incremental Delta-Bar-Delta (NOSID), an AutoStep variation \citep{dabney_adaptive_2014}, and meta-trace \citep{young_metatrace_2018}. Both SID and NOSID use a single, shared, global step size; as a result, neither are learning feature relevance. Although SID and NOSID learn step sizes using IDBD, the step sizes are not assigned on a per-feature basis. \citeauthor{dabney_adaptive_2014}'s AutoStep uses a 

In Section \ref{prediction task} We compare our methods against SID, NOSID, and AlphaBound \citep{dabney_adaptive_2012} as a sanity-check to ensure bias change and representation learning provide some benefit over methods without representation learning. In addition, we compare against \citep{dabney_adaptive_2014}'s AutoStep.

IDBD was originally derived as meta gradient-descent for Least Means Square rule learning. IDBD minimizes $\frac{\partial \delta^2_{(t)}}{\partial \beta_i(t)}$ where $\delta$ is LMS error and $\beta$ are the meta-weights such that $\exp(\beta) = \alpha$. To generalize IDBD to TD learning we must define what squared error we aim to minimize. 
\pagebreak
One option is to minimize the squared one-step TD error with respect to our meta-weights $\beta$. 

\begin{equation} \label{beta derivation semi}
\begin{split}
\beta_i(t+1) & = \beta_i(t) - \frac{1}{2} \theta \frac{\partial \delta^2(t)}{\partial \beta_i} \\
		 & = \beta_i(t) - \frac{1}{2} \theta \sum_j \frac{\partial \delta^2(t)}{\partial w_j(t)} \frac{\partial w_j(t)}{\partial \beta_i}
\end{split}
\end{equation}

To approximate $ \sum_j \frac{\partial \delta^2(t)}{\partial w_j(t)} \frac{\delta w_j(t)}{\delta \beta_i} $ we assume that $\frac{\partial w_j(t)}{\partial \beta_i} \approx 0$ where $i \neq j$. This approximation is fair, as the effect of changing the step size for a particular weight will predominantly be on the weight itself; effects on other weights will be nominal.
% mention hessian stuff here

\begin{equation}
\begin{split}
\beta_i(t+1) & \approx \beta_i(t) - \frac{1}{2} \theta \frac{\partial \delta^2(t)}{\partial w_i(t)} \frac{\partial w_{(t)}}{\partial\beta_i}
\end{split}
\end{equation}

The use of TD error introduces some subtleties based on bootstrapping: the estimate of the error $\delta = R_t + V(\phi(t+1)) - V(\phi(t))$ depends on the predicted value of the future state $V(\phi(t+1))$, resulting in a biased gradient. $R_{t+1} + \gamma V(\phi(t+1))- V(\phi(t))$ is a biased estimate of the expected return from time-step t, as it relies on on the estimate produced by the weight vector $w_t$. Because of this bootstrapping, we are not using true gradient descent \citep{barnard_temporal-difference_1993}. 

We have two choices: performing gradient descent using the full, biased gradient, or using a semi-gradient method. Semi-gradient methods do not use the estimate of the return at state $\phi(t+1)$ in the the error. In this section, we show the derivation for both choices. In the following section, we evaluate the performance of each method.

\subsection*{Semi-gradient Derivation}
For semi-gradient TD$(\lambda)$ we take the gradient of the approximate value function $V$ with respect to our weight vector $w$. In the linear case, the semi-gradient for TD($\lambda$) is simply $-\phi(t)$. Using this gradient we can find~(\ref{td gradient}) which may then be substituted back into~(\ref{beta derivation semi}).

\begin{equation} \label{td gradient}
\begin{split}
- \frac{1}{2} \frac{\partial \delta^2(t)}{\partial w_i(t)} & = - \delta(t) \frac{\partial \delta(t)}{\partial w_i(t)} \\
&= - \delta(t) \frac{\partial}{\partial w_i(t)}[- V(\phi(t))] \\
&= \delta(t) \phi(t) \\
\end{split}
\end{equation}

We then complete $\beta$'s definition in~(\ref{beta done}) by defining an additional memory vector $h$. We define $h$ as $\frac{\partial w_i(t+1)}{\partial \beta_i}$.

\begin{equation} \label{beta done}
\begin{split}
\beta_i(t+1) \approx \beta_i(t) + \theta \delta(t)\phi_i(t)h_i(t)
\end{split}
\end{equation}

We simplify $\frac{\partial w_i(t)}{\partial \beta_i}$ by describing it in terms of it's update.

\begin{equation} \label{h partial}
\begin{split}
h_i(t+1) & = \frac{\partial w_i(t+1)}{\partial \beta_i} \\
	   	 & = \frac{\partial}{\partial \beta_i} [w_i(t) + e^{\beta_i(t+1)} \delta(t) z_i(t)] \\
         & = h_i(t) + e^{\beta_i(t+1)}\delta(t)z_i(t)+e^{\beta_i(t+1)}\frac{\partial \delta(t)}{\partial \beta_i} z_i(t) +e^{\beta_i(t+1)}\frac{\partial z_i(t)}{\partial \beta_i} \delta_i(t) \\
\end{split}
\end{equation}

using the product rule to simplify (\ref{h partial}) leaves us with the remaining $\frac{\partial \delta(t)}{\partial \beta}$.

\begin{equation} \label{eq1}
\begin{split}
\frac{\partial \delta(t)}{\partial \beta_i} & = - \frac{\partial}{\partial \beta_i}[V(\phi(t))] \\
											& = - \frac{\partial}{\partial \beta_i}\sum_j w_j(t)\phi_j(t) \\
                                            & \approx - \frac{\partial}{\partial\beta_i}[w_i(t)\phi_i(t)] = -h_i(t) \phi_i(t)
\end{split}
\end{equation}

Again, as we presume that a change in step size for a particular weight will have a nominal impact on other weights, we approximate $\frac{\partial V_i(t)}{\partial \beta_i}$ as $\frac{\partial}{\partial \beta}[w_i(t)^\top \phi(t)]$. This results in $-h(t)\phi(t)$ which we may then use to simplify (\ref{h partial}) to the definition of $h$ in (\ref{h done semi}).

\begin{equation} \label{z}
\begin{split}
\frac{\partial z_i(t+1)}{\partial \beta_i} = \frac{\partial}{\partial \beta_i}[\gamma \lambda z_i(t) + \phi_i(s_{t})] = \frac{\partial z_i(t)\gamma \lambda}{\partial \beta_i} = 0
\end{split}
\end{equation}

We see that (\ref{z}) results in a decaying trace of the initialized value of the eligibility traces. Since the gradient is 0, this value will always be 0.

\begin{equation} \label{h done semi}
\begin{split}
h_i(t+1) 	& \approx h_i(t) + e^{\beta_i(t+1)}\delta(t)z_i(t) - e^{\beta_i(t+1)}\phi_i(t)z_i(t)h_i(t)\\
			& = h_i(t)[1-\alpha(t+1)\phi_i(t) z_i(t)] + \alpha_i(t+1)\delta(t)z_i(t)
\end{split}
\end{equation}

% PRE-Patrick: After positively bounding $[1-\alpha(t+1)\phi^2_i(t)]$ it is clear that the update for $h$ is as shown in Algorithm ~\ref{TIDBD_alg} , that the update for $\beta$ is as described in, and that TIDBD is a form of stochastic gradient descent for the parameter $\beta$ which updates our step size.

After positively bounding $[1-\alpha(t+1)\phi_i(t) z_i(t)]$, denoted with $+$, we have completed semi-gradient TIDBD, as shown in algorithm \ref{TIDBD_alg_semi}.

\begin{algorithm}[ht]
\caption{TIDBD($\lambda$) with semi-gradient}
\begin{algorithmic}[1]
\STATE Initialize vectors $h \in {0}^{n}$, $z \in {0}^{n} $, and both $w \in \mathbb{R}^{n}$ and $\beta \in \mathbb{R}^{n}$ as desired; initialize a scalar $\theta$; observe state $S$
% \STATE 
\STATE Repeat for each observation $s^{\prime}$ and reward $R$: 
  \INDSTATE[1] $\delta \gets R + \gamma w^\top \phi(s^{\prime}) - w^\top \phi(s)$
  \INDSTATE[1] For element $i = 1, 2, \cdots, n $: 
  	\INDSTATE[2] $\beta_i \gets \beta_i + \theta \delta \phi_i(s) h_i$
    \INDSTATE[2] $\alpha_i \gets e^{\beta_i}$
  	\INDSTATE[2] $z_i \gets z_i  \gamma \lambda + \phi_i(s)$
  	\INDSTATE[2] $w_i \gets w_i + \alpha_i \delta z_i$   % If not using traces, \phi(s_t)$
  	\INDSTATE[2] $h_i \gets h_i [1-\alpha_i\phi_i(s) z_i]^+ + \alpha_i\delta z_i$ 
  \INDSTATE[1] $s \gets s^{\prime}$
\end{algorithmic}
\label{TIDBD_alg_semi}
\end{algorithm}

One may note that semi-gradient TIDBD is similar to the original IDBD formulation (Algorithm \ref{IDBD}). The meta-weights are updated using the product of the current error $\delta$, a trace of recent weight updates $h$, and the currently active features $\phi(s)$. The most notable change is that TIDBD's $h$ trace is now modulated by not just the active features $\phi$, but also the eligibility traces $z$. This means that while the updates to step sizes will be limited to currently active features $\phi(s)$, the trace of recent weight updates include discounted past activations in $z$.

\subsection*{Ordinary Gradient Derivation}

We now derive TIDBD as stochastic meta-descent using the ordinary gradient. We start the derivation of TIDBD by describing the update rule for $\beta$---the meta-weights with which we define our step size.

Instead of using the semi-gradient, we consider both the estimated value of the state, and the target.
%pg 196/197
% by 9.8 in sutton and barto
% todo: discuss ordinary gradient
\begin{equation} \label{beta derivation}
\begin{split}
\beta_i(t+1) & \approx \beta_i(t) - \theta \delta(t) \frac{\partial \delta(t)}{\partial w_i(t)} \frac{\partial w_i(t)}{\partial\beta_i} \\
& = \beta_i(t) - \theta \delta(t) \frac{\partial[R + \gamma w^\top \phi(t+1) - w^\top \phi(t)]}{\partial w_i(t)} \frac{\partial w_i(t)}{\partial\beta_i} \\
& = \beta_i(t) - \theta \delta(t)[\gamma \phi(t+1) - \phi(t)] \frac{\partial w_i(t)}{\partial\beta_i} \\
& =\beta_i(t) - \theta \delta(t)[\gamma \phi(t+1) - \phi(t)]h_i(t)
\end{split}
\end{equation}

We then complete the simplification of $\beta$'s update by defining an additional memory vector $h$ as $\frac{\partial w}{\partial \beta}$. We then complete the update for $h$.

\begin{equation} \label{h_partial}
\begin{split}
h_i(t+1) & = \frac{\partial w_i(t+1)}{\partial \beta_i} \\
	   	 & = \frac{\partial [w_i(t) + e^{\beta_i(t+1)} \delta(t) z_i(t)]}{\partial \beta_i} \\
         & = h_i(t) + e^{\beta_i(t+1)}\delta(t)z_i(t)+e^{\beta_i(t+1)}\frac{\partial \delta(t)}{\partial \beta_i} z_i(t) + e^{\beta_i(t+1)}\frac{\partial z_i(t)}{\partial \beta_i} \delta_i(t)
\end{split}
\end{equation}

This simplification leaves us with $\frac{\partial \delta(t)}{\partial \beta_i}$, derived in (\ref{eq1}), and $\frac{\partial z_i(t)}{\partial \beta_i}$, simplified in (\ref{z}). We use the same approximation as in (\ref{beta derivation semi}) to simplify:

\begin{equation}
\begin{split}
\frac{\partial \delta(t)}{\partial \beta_i} & =\frac{\partial}{\partial \beta_i}[R + \gamma w^\top \phi(t+1) - w^\top \phi(t)] \\
											& =\frac{\partial}{\partial \beta_i} [\sum_j R + \gamma w_j \phi_j(t+1) - w_j \phi_j(t)] \\
                                            & \approx \frac{\partial}{\partial\beta_i}[R + \gamma w_i \phi_i(t+1) - w_i \phi_i(t)] \\
                                            & = \gamma h_i \phi_i(t+1) - h_i \phi_i(t)
\end{split}
\end{equation}

%Again, we approximate $\frac{\partial \sum_j w_j(t)\phi_j(t)}{\partial \beta_i}$ as $\frac{\partial w_i(t) \phi_i(t)}{\partial \beta_i}$.
\begin{equation} \label{h done}
\begin{split}
h_i(t+1) 	& \approx h_i(t) + e^{\beta_i(t+1)}\delta(t)z_i(t) + e^{\beta_i(t+1)} [\gamma h_i \phi_i(s_{t+1}) - h_i \phi_i(s_{t})] z_i(t)  + 0 e^{\beta_i(t+1)} \delta_i(t)\\ 
			& = h_i(t)[1+\alpha_i(t+1) z_i(t)[\gamma \phi_i(s_{t+1}) - \phi_i(s_{t})]] + \alpha_i(t+1)\delta(t)z_i(t)
\end{split}
\end{equation}

We then take the results from (\ref{eq1}) and (\ref{z}) to complete the definition of $h$'s update. The update for $h$ and $\beta$ may then be implemented directly as shown earlier in Algorithm \ref{TIDBD_alg_full}.

%IDBD originally suggested that the step sizes be capped to prevent underflow, and that the update to a feature be limited between $\pm 2$ to prevent radical changes to the value of the step sizes. I chose not to add these constraints, as they are unprincipled modifications; they cover-up some of the limitations of IDBD without actually fixing them. Here, I attempt to fix these problems and create a complete and stable version of TIDBD.

\begin{algorithm}[tbhp!]
\caption{TIDBD($\lambda$)}
\begin{algorithmic}[1]
\STATE Initialize vectors $h \in {0}^{n}$, $z \in {0}^{n} $, and both $w \in \mathbb{R}^{n}$ and $\beta \in \mathbb{R}^{n}$ as desired; initialize a scalar $\theta$; observe state $S$
% \STATE 
\STATE Repeat for each observation $s^{\prime}$ and reward $R$: 
  \INDSTATE[1] $\delta \gets R + \gamma w^\top \phi(s^{\prime}) - w^\top \phi(s)$
  \INDSTATE[1] For element $i = 1, 2, \cdots, n $: 
  	\INDSTATE[2] $\beta_i \gets \beta_i - \theta \delta [\gamma \phi(s^\prime) - \phi(s)] h_i$
    \INDSTATE[2] $\alpha_i \gets e^{\beta_i}$
  	\INDSTATE[2] $z_i \gets z_i  \gamma \lambda + \phi_i(s)$
  	\INDSTATE[2] $w_i \gets w_i + \alpha_i \delta z_i$   % If not using traces, \phi(s_t)$
  	\INDSTATE[2] $h_i \gets h_i[1+\alpha_i z_i[\gamma \phi_i(s^\prime) - \phi_i(s)]]^+ + \alpha_i \delta z_i$ 
  \INDSTATE[1] $s \gets s^{\prime}$
\end{algorithmic}
\label{TIDBD_alg_full}
\end{algorithm}

%The most notable difference between an ordinary gradient TIDBD and a semi-gradient TIDBD is the change from $-\phi$ to $\gamma \phi(s^\prime)-\phi(s)$. % todo: add analysis here: what does this mean. 

\section{Does TIDBD(0) With a Single, Shared step size Outperform Ordinary TD?}

Having derived both ordinary and semi-gradient TIDBD, we now evaluate whether the benefits of IDBD transfer to TD learning. 

First, we assess the ability of TIDBD to improve upon traditional TD prediction in a simple tabular setting. In a tabular setting, the advantages of vectorizing step sizes are not present, enabling us to assess whether adapting step sizes with TIDBD is an improvement over ordinary TD in general, independent of performing representation learning. 

An ideal bias learning method would be able to perform as well as or better than TD for arbitrary initial step sizes  $\alpha_0$ while being insensitive to meta-parameters. We expect that TIDBD should be able to out-perform TD for all $\theta$ values, and that it should be relatively insensitive to the choice of $\theta$ values.

\subsection{Gridworld}
\label{gridworld-problem}

\begin{figure}
\centering
\includegraphics[width=\linewidth, height=2in, keepaspectratio]{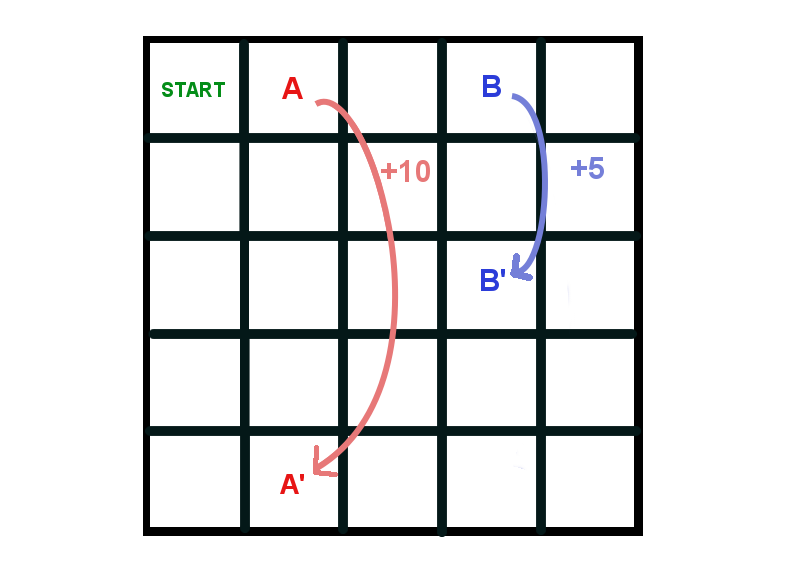}
\caption{Gridworld Problem as introduced in \citep{sutton_reinforcement_1998}.}
\label{gridworld}
\end{figure}

We created a suitable prediction task by generating a Markov Reward Process from a grid-world problem originally described in \cite{sutton_reinforcement_1998} (depicted in Figure \ref{gridworld}). Each tile in the 5 $\times$ 5 grid-world represents a state. The state transitions are the four cardinal directions---north, south, east, and west---chosen by an equiprobable random policy. Transitions which would leave the grid resulted in staying in the same state and a reward of -1. Regardless of the transition in state $A$ or $B$, the learner transitions to states $A^\prime$ and $B^\prime$ respectively with a probability of 1. A transition from $A$ to $A^\prime$ yields a reward of 10 and a transition from $B$ to $B^\prime$ yields a reward of 5. All other transitions receive a reward of 0. The start state was the top left-hand corner. A trial consisted of the equiprobable random policy acting for 15000 time-steps. Each prediction method would then learn a value function over the 30 trials.

We compared ordinary TD to both semi-gradient and ordinary gradient TIDBD with initial step sizes distributed between 0.0005 and 0.5. For all prediction methods $\lambda = 0$ and $\gamma = 0.99$. We swept over 21 different meta-parameters equally distributed between the range of $0 < \theta < 0.2$---the range for which IDBD was originally compared over in \cite{sutton_adapting_1992}. When $\theta = 0$, TIDBD and TD are equivalent, as the initial step size $\alpha_0$ is never updated.

\begin{figure*}[tbhp!]
\centering
\begin{subfigure}{\textwidth}
    \centering
    \includegraphics[width=\linewidth, keepaspectratio]{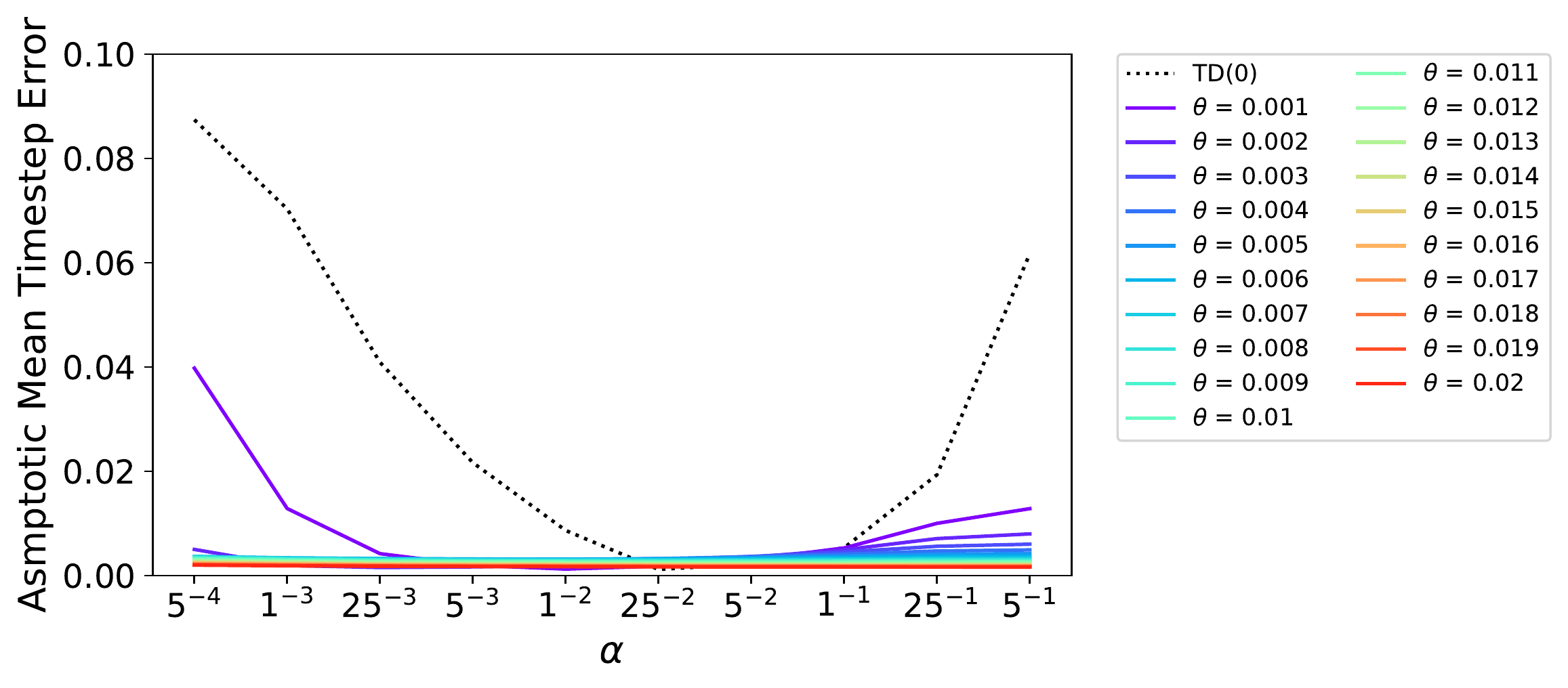}
\caption{Semi-gradient TIDBD.}
\label{error_}
\end{subfigure}
\begin{subfigure}{\textwidth}
	\centering
    \includegraphics[width=\linewidth, keepaspectratio]{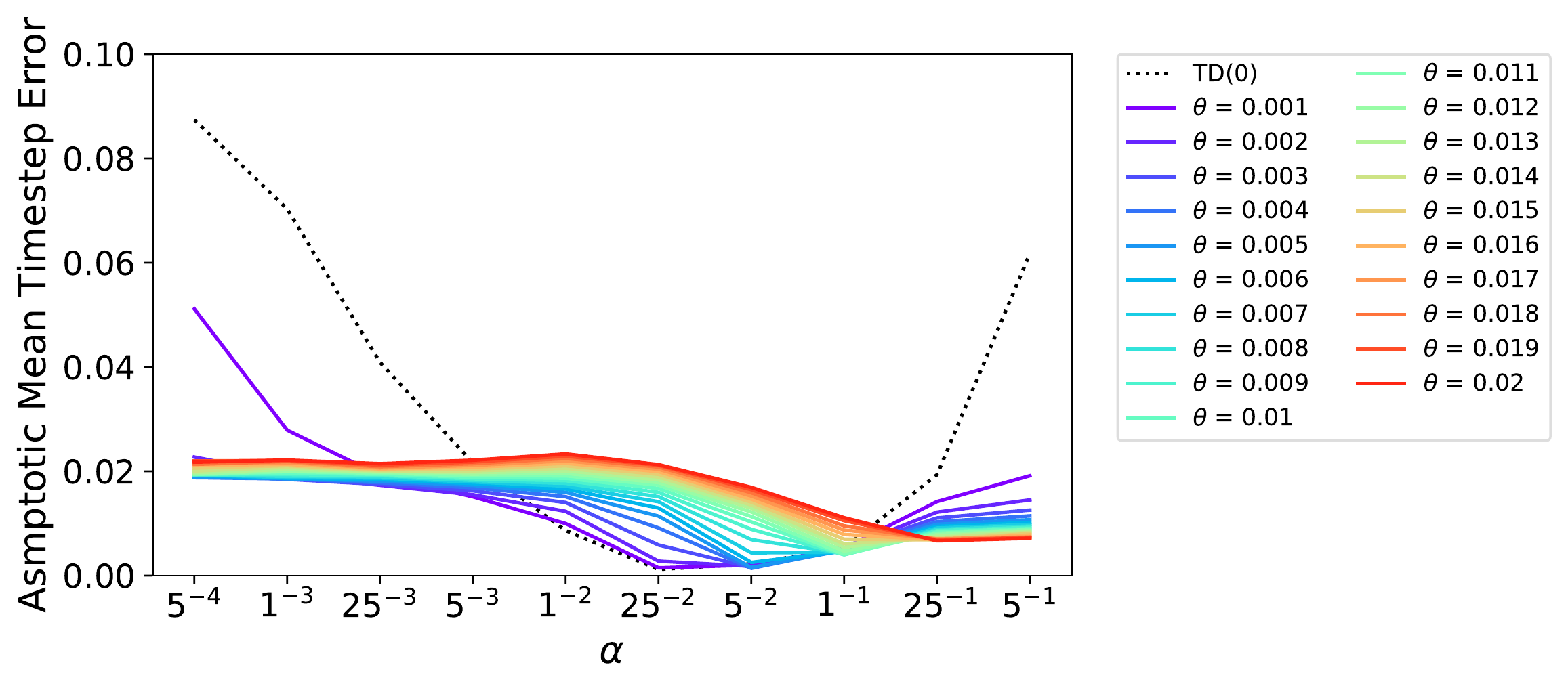}
\caption{Ordinary gradient TIDBD.}
\label{ordinary_gradient_gridworld}
\end{subfigure}
\caption{Parameter study of semi and ordinary gradient TIDBD. Static step size TD in black.}
\label{parameters}
\end{figure*}

Figure \ref{parameters} depicts the performance of both semi-gradient and ordinary gradient TIDBD for settings of their meta step size. As expected, semi-gradient TIDBD is less sensitive to meta step size $\theta$ values than ordinary TIDBD, but has a higher asymptotic error than ordinary TD for the best initial step size setting $\alpha_0 = 0.05$. For all but $\alpha_0 = 0.05$, there are broad settings of $\theta$ such that TIDBD attains better asymptotic performance than ordinary TD. As we move further away from the optimal step size, the greater the advantage of adapting the step sizes with TIDBD becomes.

The semi-gradient accounts for the effect of changing the weights on the estimate, but ignores the effects on the target. While semi-gradient methods do not converge as robustly, they converge fast with reliability. The motivation for a semi-gradient TIDBD is to make the method less sensitive to the setting of $\theta$. By not taking into account the target, non-stationarity in the update of the target value will not affect the updates of, making TIDBD more stable than an ordinary gradient TD method.

% for every initial step size, using TIDBD improves upon ordinary fixed-step size TD
For every initial step size except for the best setting ($\alpha_0 = 0.05$), there are settings of $\theta$ such that TIDBD is an improvement over ordinary fixed step size TD. Common to both ordinary and semi-gradient TIDBD is the asymmetry of sensitivity to meta-parameters: the performance of TIDBD is more robust for initializations of $\alpha > 0.05$. In general, TIDBD is more robust to larger initial step size settings than it is to smaller initial step size settings. This is intuitive. Step size schedules typically start with large values which decrease over time. 

In figure \ref{ordinary_gradient_gridworld}, the performance of TIDBD with an ordinary gradient is displayed across varying settings of $\alpha$ and $\theta$. As anticipated, TIDBD with an ordinary gradient is more sensitive to $\theta$ settings.Unlike semi-gradient TIDBD, TIDBD with an ordinary TD gradient achieves performance equivalent to ordinary TD for the best initial step size.

\begin{figure}[!htbp] 
\centering
% \begin{minipage}[t]{\dimexpr.45\textwidth-.5\columnsep}

\begin{subfigure}{\textwidth}
\centering
\includegraphics[height=2in, width=\linewidth, keepaspectratio]{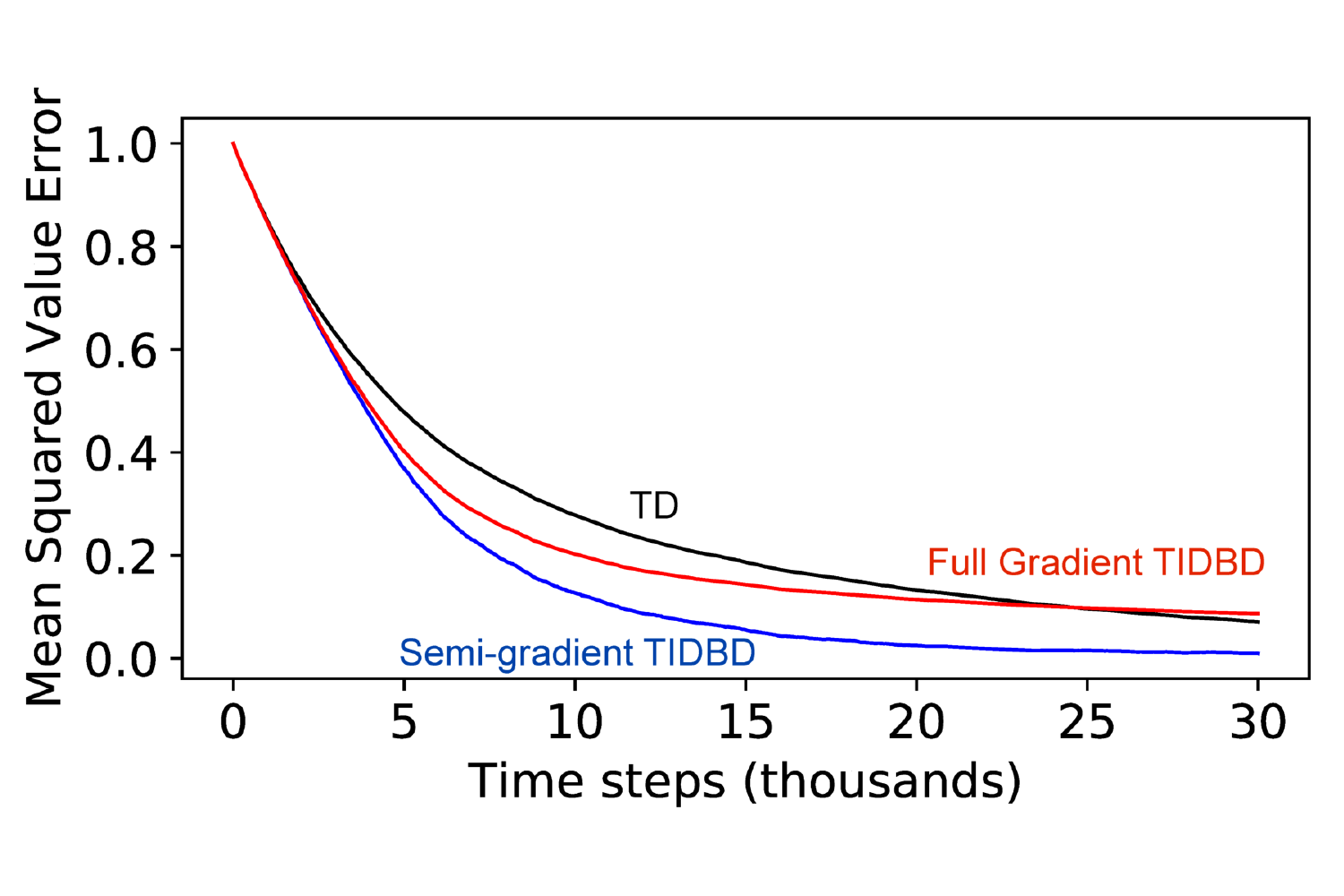}
\caption{$\alpha_0 = 0.01$}
\label{error_1}
\end{subfigure}

\begin{subfigure}{\textwidth}
\centering
\includegraphics[height=2in, width=\linewidth,keepaspectratio]{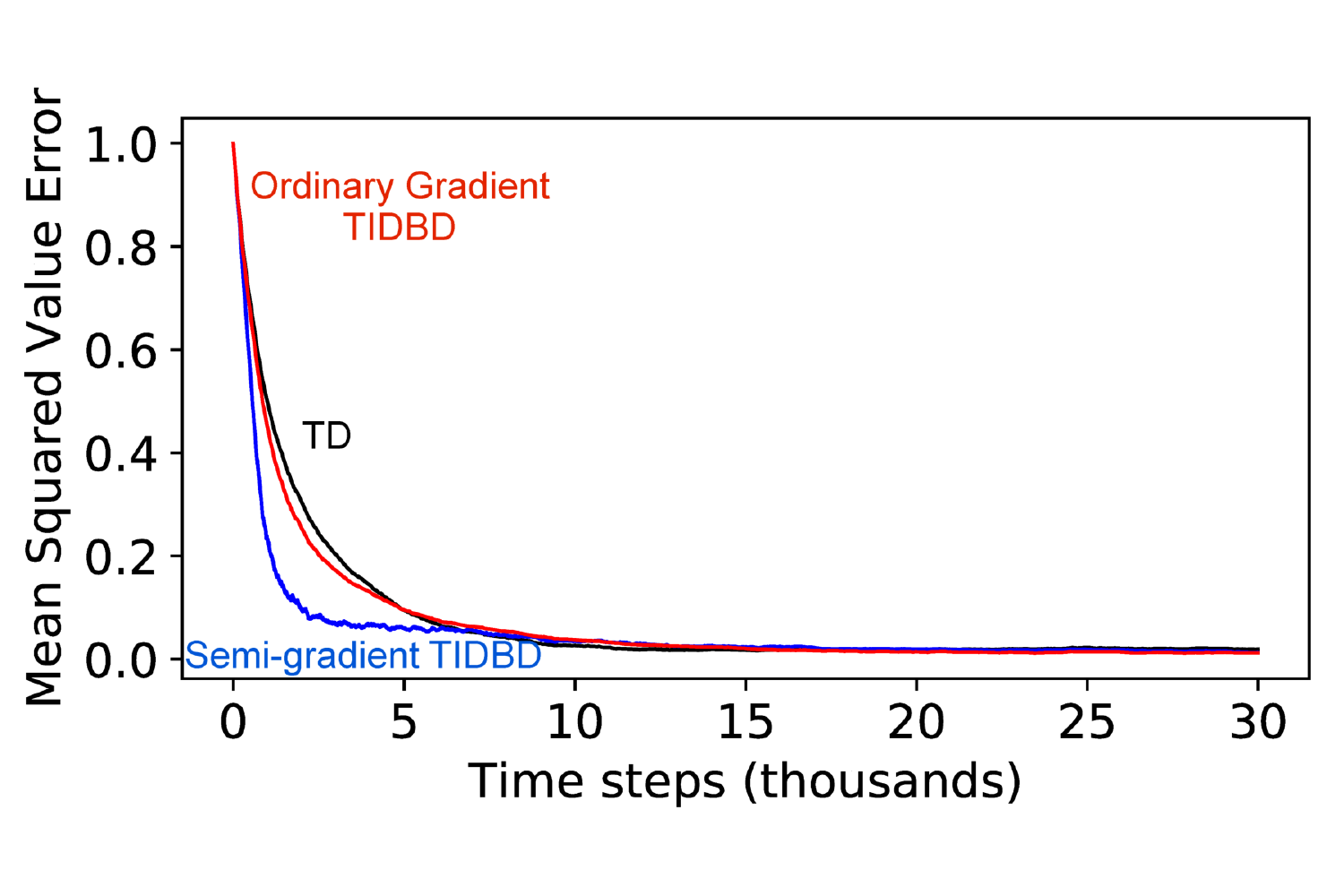}
\caption{$\alpha_0 = 0.05$}
\label{error_2}
\end{subfigure}

\begin{subfigure}{\textwidth}
\centering
\includegraphics[height=2in, width=\linewidth, keepaspectratio,]{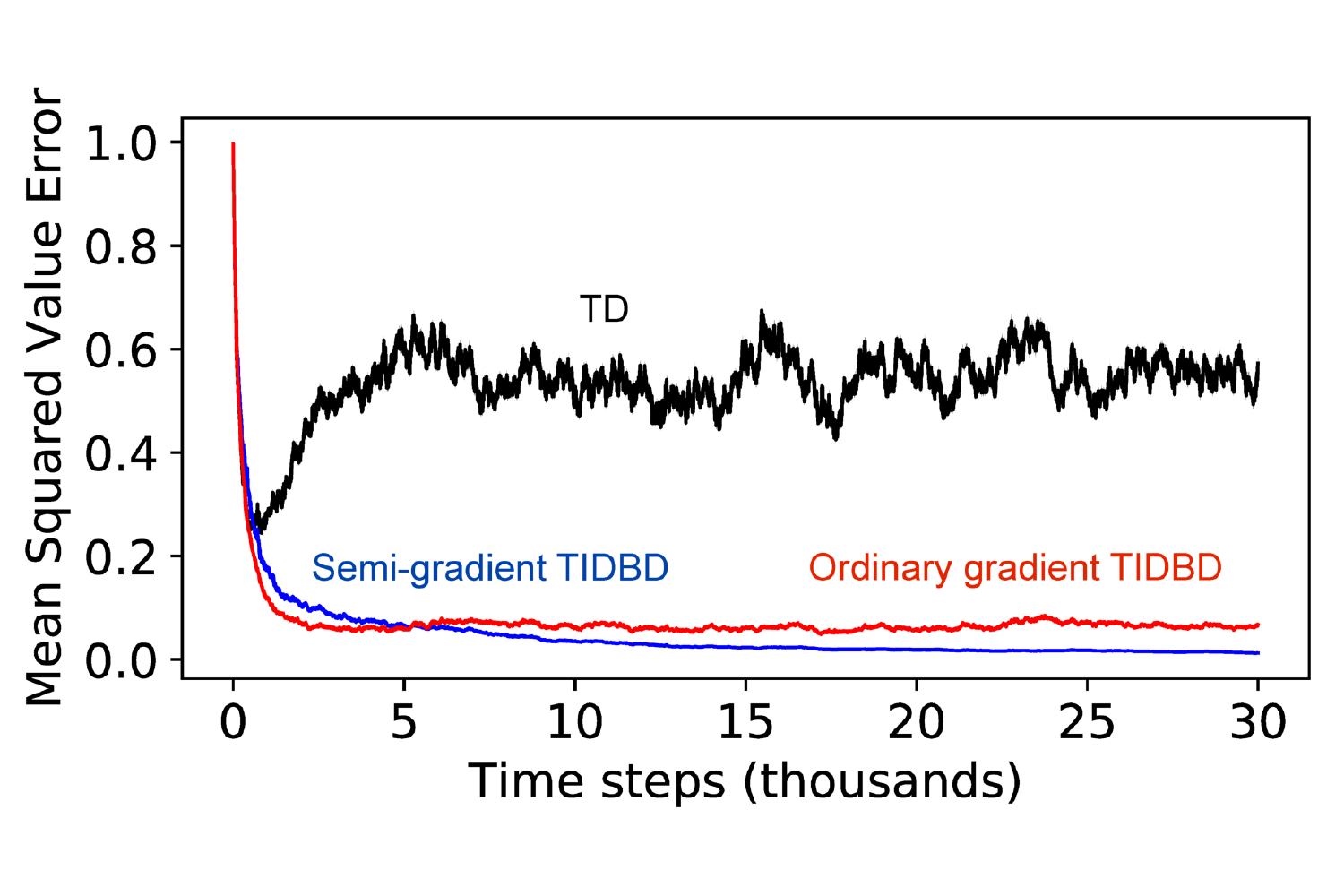}
\caption{$\alpha_0 = 0.5$}
\label{error_3}
\end{subfigure}

\caption{Error during learning for three initializations $\alpha_0$ for TD(0) and both semi-gradient TIDBD and ordinary-gradient TIDBD.}
\label{initial_learning_curves}
% \end{minipage}%
\end{figure}

% re-write these sections

In Figure \ref{initial_learning_curves} the error of both semi-gradient and ordinary TIDBD are compared to TD for three initial step size settings. Figure \ref{error_1} is the most conservative initialization of $\alpha_0$. We expect that for this $\alpha_0$ learning for ordinary TD will be slow, as it takes small steps when updating it's estimated value for each state. For $\alpha_0 = 0.01$, both ordinary and semi-gradient TIDBD outperform ordinary TD, with both increasing their step sizes enabling them to make larger updates to their weights and learn more quickly.

For the best setting $\alpha_0 = 0.05 ($Figure \ref{error_2}), there is little asymptotic difference between TD both versions of TIDBD; however, semi-gradient TIDBD learns far faster than both ordinary TD and ordinary gradient TIDBD. In early learning, semi-gradient TIDBD is able to find step sizes which enable it to converge the fastest. The benefits of both semi and ordinary-gradient TIDBD are most apparent at the most aggressive setting of $\alpha_0 = 0.5$ (Figure \ref{error_3}). While ordinary TD is unable to learn, both semi and ordinary gradient TIDBD are able to learn in spite of the poor initial step size setting---they are able to tune their step sizes down in response. 

%todo: sum this up, brah

% for optimal initial step sizes, TIDBD has both lower error than.

% The benefit of reduced variance is made more clear in the highest step size. 

% for this problem, the best tuned meta step size is consistent across all initial alphas

% From the shape of the error curve, we can see that the benefit of TIDBD is not consistent across all initial step sizes. For larger step sizes, the benefits of IDBD are more pronounced for the range of meta step sizes we have chosen. 

\section{AutoTIDBD: AutoStep-Style TIDBD}

TIDBD with an ordinary gradient  was able to perform as well as or better than TD for some $\theta$ value at each $\alpha_0$ value; however, the best $\theta$ values varied for different $\alpha_0$ values. Conversely, semi-gradient TIDBD had broad ranges of values of $\theta$ for which the performance was acceptable, but did not perform as well as TD for the best value of $\alpha_0$.

One of the benefits of a static shared step size is that most reinforcement learning practitioners have an intuition of what range of values will yield acceptable performance in general; however, the optimal step size value will vary from domain to domain. With TIDBD and other similar IDBD methods, the performance is dependent on a meta step size---a value for which there is little intuition as to how to set. A poor choice of meta step size can lead to explosive updates.

%Upon closer inspection, the origin of the divergence seemed to lie with the un-normalized nature of the $\beta$ weight updates. For the base learning algorithm, we normalize the inputs and use them to construct a binary feature vector. The weights are normalized so that they're proportional and all in the same range.
% todo: expand

The instability of IDBD is not a new observation. \cite{sutton_adapting_1992} originally suggested bounding both step size values and the size of updates to the meta-weights $\beta$---amendments we chose to exclude so as to better understand TIDBD's underlying performance. AutoStep---an extension of IDBD---reduced sensitivity to the setting of the meta step size and prevented divergence in the supervised learning setting\citep{mahmood_tuning-free_2012}. To prevent divergence, AutoStep makes two additions to IDBD. First, the meta-weight update is normalized by a decaying average of recent weight updates. Second, the step sizes is normalized by the amount by which the error was reduced on a given example---termed the \textit{effective step size}. By normalizing the current step sizes by the effective step size a weight update will never overshoot on the current observed example.

% this could be far better.

\subsection{AutoStep}

To manage explosive growth of step sizes, Autostep (Algorithm \ref{autostep}) adds two components to IDBD (Algorithm \ref{IDBD}): a normalization of the inputs $\delta x h$ with which we update our $\beta$ meta-weights, and a normalization of the resulting step size $\exp(\beta)$ by the \textit{effective step size}. 

% todo: explain the decay

To normalize the inputs, AutoStep maintains $\eta$, a running trace of recent weight updates (Line 5). At each time-step it takes the maximum between the current weight update $|\delta x h|$ and a decay of the previous maximum $v_i + \frac{1}{\tau} \alpha_i x_i^2 (|\delta x_i h_i|-\eta)$. The scalar $\tau$ is a large value which weights the decay of $v$. 

\begin{equation}
\eta \gets \max(|\delta x h|, \eta + \frac{1}{\tau} \alpha x^2 (|\delta x h| - \eta))
\end{equation}

One might consider why the maximum is decayed rather than simply stored---as is done with NOSID (see \cite{dabney_adaptive_2014} for further details). In real-world data sources, noise and other outliers could distort the absolute recorded maximum, making the normalizer adjust input values into an unrepresentative range. By decaying the maximum, we enable the system to recover gracefully and gradually from such extreme data.

On line 7, this normalizer $\eta$ is used to make the update $\delta x h$ unitless.

After the meta-weights have been updated, the resulting step size $\alpha$ is normalized by the \textit{effective step size} (line 8). The effective step size describes the amount by which we reduce the error on the current example by making a weight update. An effective step size equal to one means that the error has been entirely reduced for the current example. By dividing the current step size $\alpha$ by $\max(1, \textit{effective step size})$, we prevent over-shooting on a given example: we prevent an update which introduces more error.

%AutoStep's changes to IDBD make it less sensitive to the choice of $\theta$, although the reasons for this are not known, and it is not clear to what extent this generalizes. In addition to reduced sensitivity to the meta step size $\theta$, AutoStep is also less sensitive to it's initial step size setting for $\beta$. Continuing this trend of relative insensitivity, $\tau$. While it adds a parameter to IDBD, AutoStep behaves like a parameter-free method, making it an ideal starting point for stabilizing TIDBD.

\begin{algorithm}[htbp!]
\caption{AutoStep}
\begin{algorithmic}[1]
\STATE Initialize vectors $h \in {0}^{n}$, $z \in {0}^{n} $, and both $w \in \mathbb{R}^{n}$ and $\beta \in \mathbb{R}^{n}$ as desired; initialize vector $\alpha \gets \exp(\beta)$ initialize a scalar $\theta$; observe state $S$
\STATE Repeat for each observation $x$ and target $y$: 
  \INDSTATE[1] $\delta \gets y - \textbf{w}^\top x$
  \INDSTATE[1] For element $i = 1, 2, \cdots, n $: 
  	\INDSTATE[2] $\eta_i \gets \max(|\delta x_i h_i|, \eta_i + \frac{1}{\tau} \alpha_i x_i^2 (|\delta x_i h_i|-\eta))$
    \INDSTATE[2] If $\eta_i \neq 0$:
    	\INDSTATE[3] $\alpha_i \gets \alpha_i \exp(\mu \frac{\delta x_i h_i}{\eta_i})$
	\INDSTATE[1] $M \gets \max(\sum(\alpha_i x_i^2), 1)$
  \INDSTATE[1] For element $i = 1, 2, \cdots, n $: 
  	\INDSTATE[2] $\alpha_i \gets \frac{\alpha_i}{M}$
    \INDSTATE[2] $w_i \gets w_i + \alpha_i \delta x_i$
    \INDSTATE[2] $h_i (1-\alpha_i x^2_i) + \alpha_i \delta x_i$
\end{algorithmic}
\label{autostep}
\end{algorithm}

\subsection{AutoTIDBD: AutoStep for TD learning}

Having introduced AutoStep, we now add Auto-step's normalization to TIDBD to improve its stability. AutoStep prevents divergence by normalizing the current step size $\alpha_i$ by the effective step size $\alpha_i^\top x_i^2$---the amount by which the error on the current example is reduced by updating the weight vector. If the effective step size is one, then we have reduced all error on the given example; If the effective step size is greater than one, then we have over-corrected on a given example. If we divide the current step size by the effective step size before performing a weight update in this instance, we ensure that we do not overshoot on the given example.

We calculate the effective step size by taking the difference between the error before the weight update $\delta_t(t)$ and the error after the weights have been updated $\delta_{t+1}(t)$, or $\frac{\delta_t(t)-\delta_{t+1}(t)}{\delta_t(t)}$. We calculate the error  $\delta_{t+1}(t)$ using the weights from time-step $t+1$ and the observation from time-step $t$. 

For supervised learning, the notion of an effective step size is straightforward: there is a known target value, so the error reduced on a given time-step is directly observable. However, as previously mentioned, TD learning uses bootstrapping. For TD learning the effective step size is not an exact value, but a biased estimation dependent on how accurate the value-function is in estimating the value of the following step.

% TODO: justifyyyyy
\begin{equation} \label{TDESS_A}
\begin{split}
 \frac{\delta_t(t) - \delta_{t+1}(t)}{\delta_t(t)} & = \frac{[R_{t+1} + \gamma V_t(\phi(t+1) - V_t(\phi(t))] - [R_{t+1} + \gamma V_{t+1}(\phi(t+1) - V_{t+1}(\phi(t))]}{\delta(t)} \\ 
 &=  \frac{[\gamma V_t(\phi(t+1) - V_t(\phi(t))]}{\delta(t)}  \\ &- \frac{[\gamma (V_{t}(\phi(t+1) + (\alpha \delta z)^\top \phi(t+1) )  - (V_{t}(\phi(t)) + (\alpha \delta z)^\top \phi(t) ]}{\delta(t)} \\
\end{split}
\end{equation}

% todo: make the notation of alpha by timestpe consistent

We expand $\delta_{t+1}(t)$, as the TD error of the current time-step $t$ using the value-functions from the following time-step, $V_{t+1}$. Value functions may be written recursively as the sum of the previous time-step's value-function $V_t(\phi(s_t)$ and the current weight update $\alpha_t \delta_t{t} z_t$. So, $V_{t+1}(\phi(s_{t})) = V_t(\phi(s_t)) + [\alpha_t \delta_t(t) z_t](\phi(s_t))$.

\begin{equation} \label{TDESS_B}
\begin{split}
 \frac{\delta_{t}(t) - \delta_{t+1}(t)}{\delta_{t}(t)}  & = \frac{[\gamma V_t(\phi(t+1) - \gamma V_{t}(\phi(t+1)]}{\delta(t)} 
 \\  & - \frac{[V_t(\phi(t)-V_{t}(\phi(t)] - [\gamma \alpha \delta z^\top  \phi(t+1) - \alpha \delta z^\top  \phi(t) ]}{\delta(t)} \\
 & = \frac{ -[\gamma \alpha \delta z^\top  \phi(t+1) )  - \alpha \delta  z^\top \phi(t) ]}{\delta(t)} \\
  & = -(\alpha z)^\top  [\gamma \phi(t+1) - \phi(t) ] \\
\end{split}
\end{equation}

% todo: elucidate the effective step size for the RL case

The resulting effective step size is $-(\alpha z)^\top  [\gamma \phi(t+1) - \phi(t) ]$. This is an intuitive result, as the amount by which we will reduce our error on a given example is the difference between the the update made to the features active in the target $\phi(t+1)$ and the changes made to the features in the state who's value we are currently estimating $\phi(t)$. 
%It's also similar, you might note, to the decay term in the update of $h$: $h_i \gets h_i [1 + \alpha_i [\gamma \phi_i(s_{t+1}) - \phi_i(s)]  z_i]^+ + \alpha_i\delta z_i$.

With the effective step size defined, what remains in defining an AutoStep for TIDBD is the weight-update's normalizing term. AutoStep simply maintains a running trace of the absolute value of the weight-updates $\max(|\delta x_i h_i|, v_i + \frac{1}{\tau} \alpha_i x_i^2 (|\delta x_i h_i|-v))$. The absolute weight update for TIDBD is $|\delta [\gamma \phi(s_{t+1}) - \phi(s)] h|$, and the current active step size is $\alpha [\gamma \phi(s_{t+1}) - \phi(s_t)]$. Thus, the trace $\eta$ of the maximum weight update would be $\max(|\delta [\gamma \phi_i(s_{t+1}) - \phi_i(s)] h_i|, \eta_i - \frac{1}{\tau} \alpha_i [\gamma \phi_i(s_{t+1}) - \phi_i(s)]  z_i(|\delta \phi_i(s) h_i| - \eta_i)) $

\begin{algorithm}[H]
\caption{AutoStep Style Normalized TIDBD($\lambda$)}
\begin{algorithmic}[1]
\STATE Initialize vectors $h \in {0}^{n}$, $z \in {0}^{n} $, and both $w \in \mathbb{R}^{n}$ and $\beta \in \mathbb{R}^{n}$ as desired; initialize a scalar $\theta$; observe state $S$
\STATE Repeat for each observation $s^{\prime}$ and reward $R$: 
  \INDSTATE[1] $\delta \gets R + \gamma w^\top \phi(s^{\prime}) - w^\top \phi(s)$
  \INDSTATE[1] For element $i = 1, 2, \cdots, n $: 
    \INDSTATE[2] $\eta_i \gets \max[ $\\ 
    \INDSTATE[4] $|\delta [\gamma \phi_i(s^{\prime}) - \phi_i(s)] h_i|,$\\
    \INDSTATE[4] $\eta_i - \frac{1}{\tau} \alpha_i [\gamma \phi_i(s^{\prime}) - \phi_i(s)]  z_i(|\delta \phi_i(s) h_i| - \eta_i)]$ \\ 
  \INDSTATE[1] For element $i = 1, 2, \cdots, n $: 
  	\INDSTATE[2] $\beta_i \gets \beta_i - \theta \frac{1}{\eta_i}\delta [\gamma \phi_i(s^{\prime})) - \phi_i(s)] h_i$
    \INDSTATE[2] $M \gets max(-e^{\beta_i} [\gamma \phi_i(s^{\prime}) - \phi_i(s)]^\top z_i$, 1)
    \INDSTATE[2] $\beta_i \gets \beta_i - \log(M)$
    \INDSTATE[2] $\alpha_i \gets e^{\beta_i}$
  	\INDSTATE[2] $z_i \gets z_i  \gamma \lambda + \phi_i(s)$
  	\INDSTATE[2] $w_i \gets w_i + \alpha_i \delta z_i$   % If not using traces, \phi(s_t)$
  	\INDSTATE[2] $h_i \gets h_i [1 + \alpha_i [\gamma \phi_i(s^{\prime}) - \phi_i(s)]  z_i]^+ + \alpha_i\delta z_i$ 
  \INDSTATE[1] $s \gets s^{\prime}$
\end{algorithmic}
\label{auto_step_tidbd_alg}
\end{algorithm}

% todo: add intuition

With the generalization to AutoTIDBD, we assess it's performance to determine whether it is able to perform better or equal to than tuned ordinary TD(0) while being relatively insensitive to its meta step size $\theta$, meeting one of our core criteria for an adaptive bias algorithm.

\pagebreak

\subsection{AutoTIDBD in Gridworld}

\begin{figure}[H]
\centering
\includegraphics[width=\linewidth,keepaspectratio]{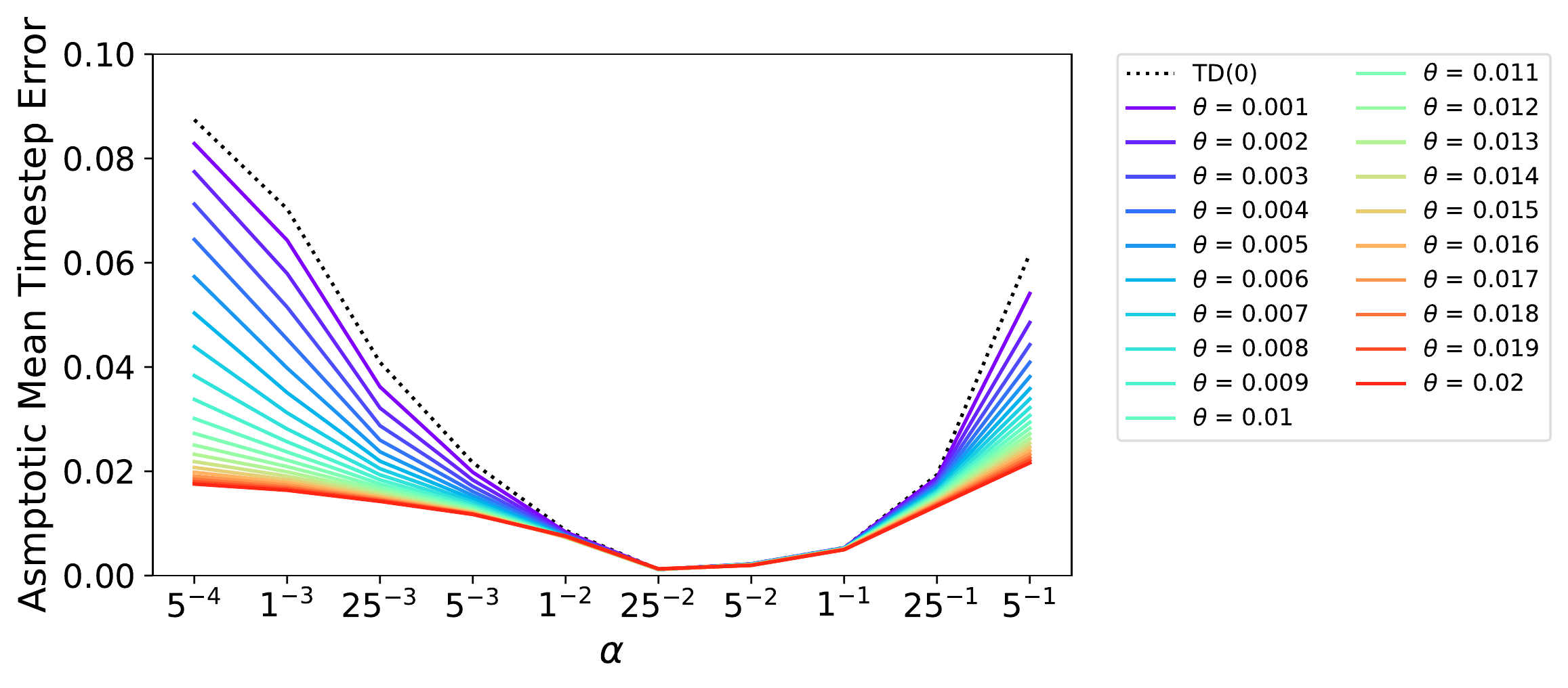}
\caption{Parameter study of Full-gradient AutoTIDBD for varying $\alpha_0$ and $\theta$ values.}
\label{gridworld_autotidbd_full}
\end{figure}

\begin{figure}[H]
\centering
\includegraphics[width=\linewidth,keepaspectratio]{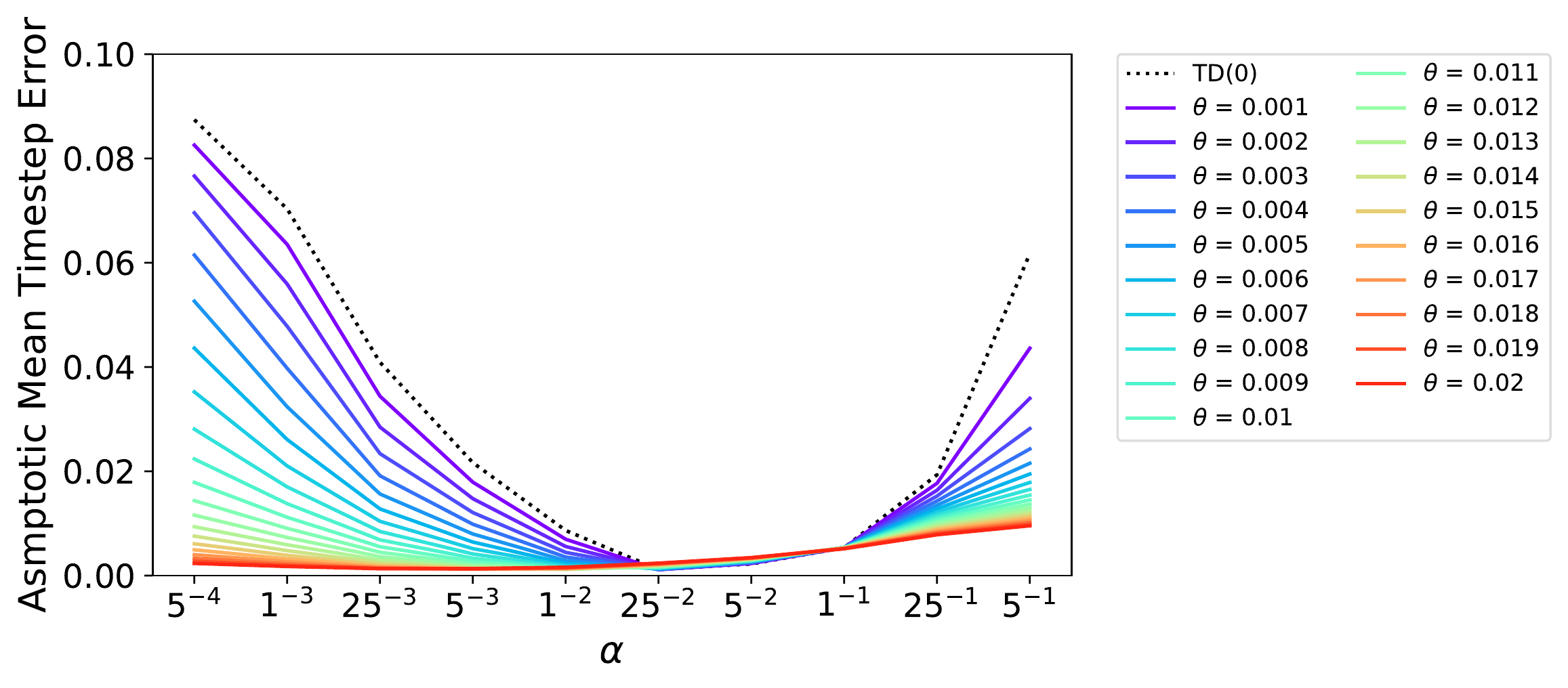}
\caption{Parameter study of semi-gradient AutoTIDBD for varying $\alpha_0$ and $\theta$ values.}
\label{gridworld_autotidbd_semi}
\end{figure}

In Figure \ref{gridworld_autotidbd_semi} and \ref{gridworld_autotidbd_full}, a parameter study of AutoTIDBD's sensitivity on the task introduced in section \ref{gridworld-problem} is presented. As with the previous experients, AutoTIDBD is adapting a single, shared step size.

Similar to semi-gradient TIDBD, there are broad ranges of $\theta$ values for which AutoTIDBD outperforms ordinary TD; like ordinary-gradient TIDBD, AutoTIDBD performs as well as or better than TD, even for the best $\alpha_0$ setting. While the absolute best performance may vary for different values of $\alpha_0$, the change in performance as $\theta$ varies is predictable and consistent.

\begin{figure}[H] 
% \begin{minipage}[t]{\dimexpr.45\textwidth-.5\columnsep}

\begin{subfigure}{\textwidth}
\centering
\includegraphics[height=2in, width=\linewidth, keepaspectratio]{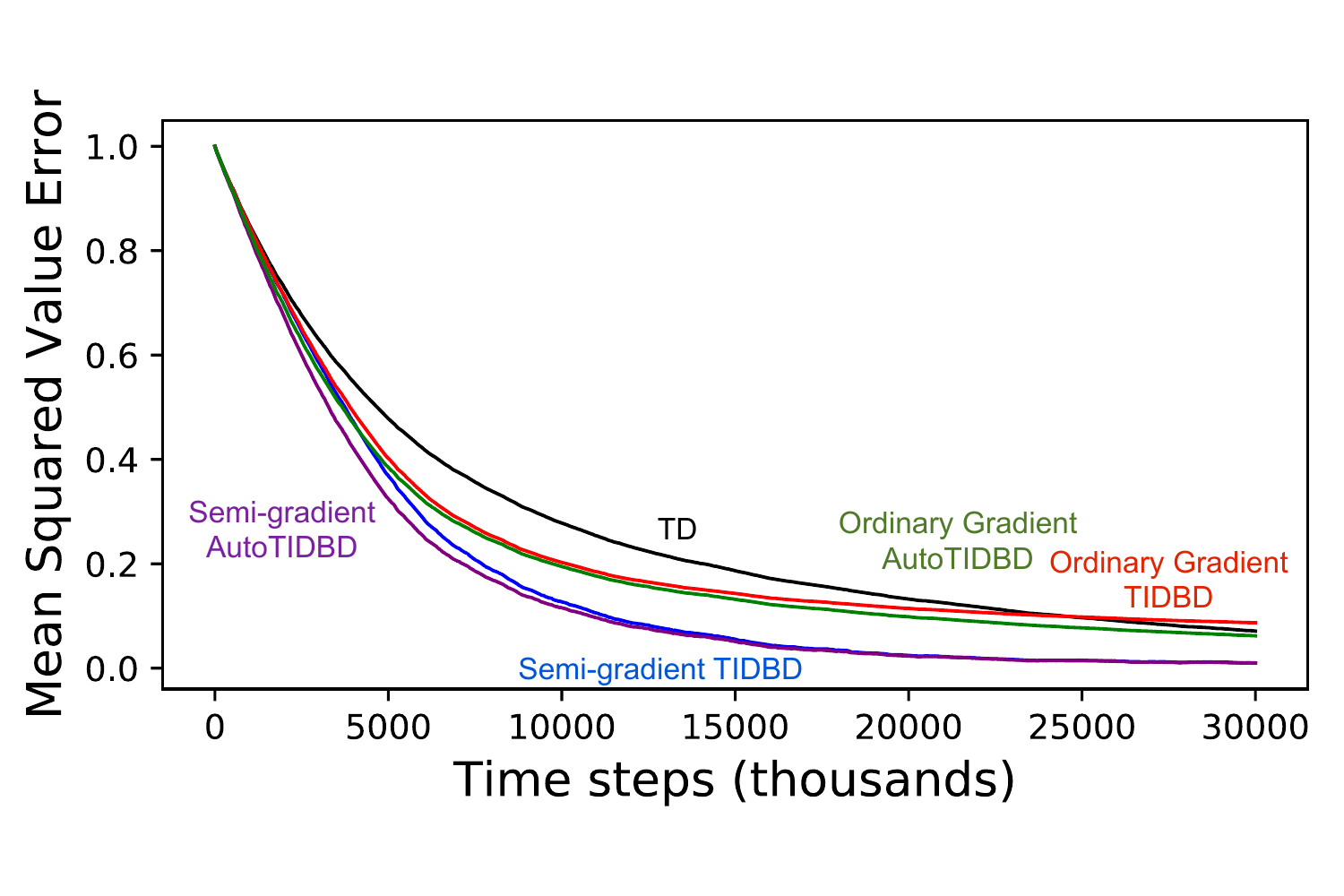}
\caption{$\alpha_0 = 0.01$}
\label{error_1_w_auto}
\end{subfigure} \\

\begin{subfigure}{\textwidth}
\centering
\includegraphics[height=2in, width=\linewidth, keepaspectratio]{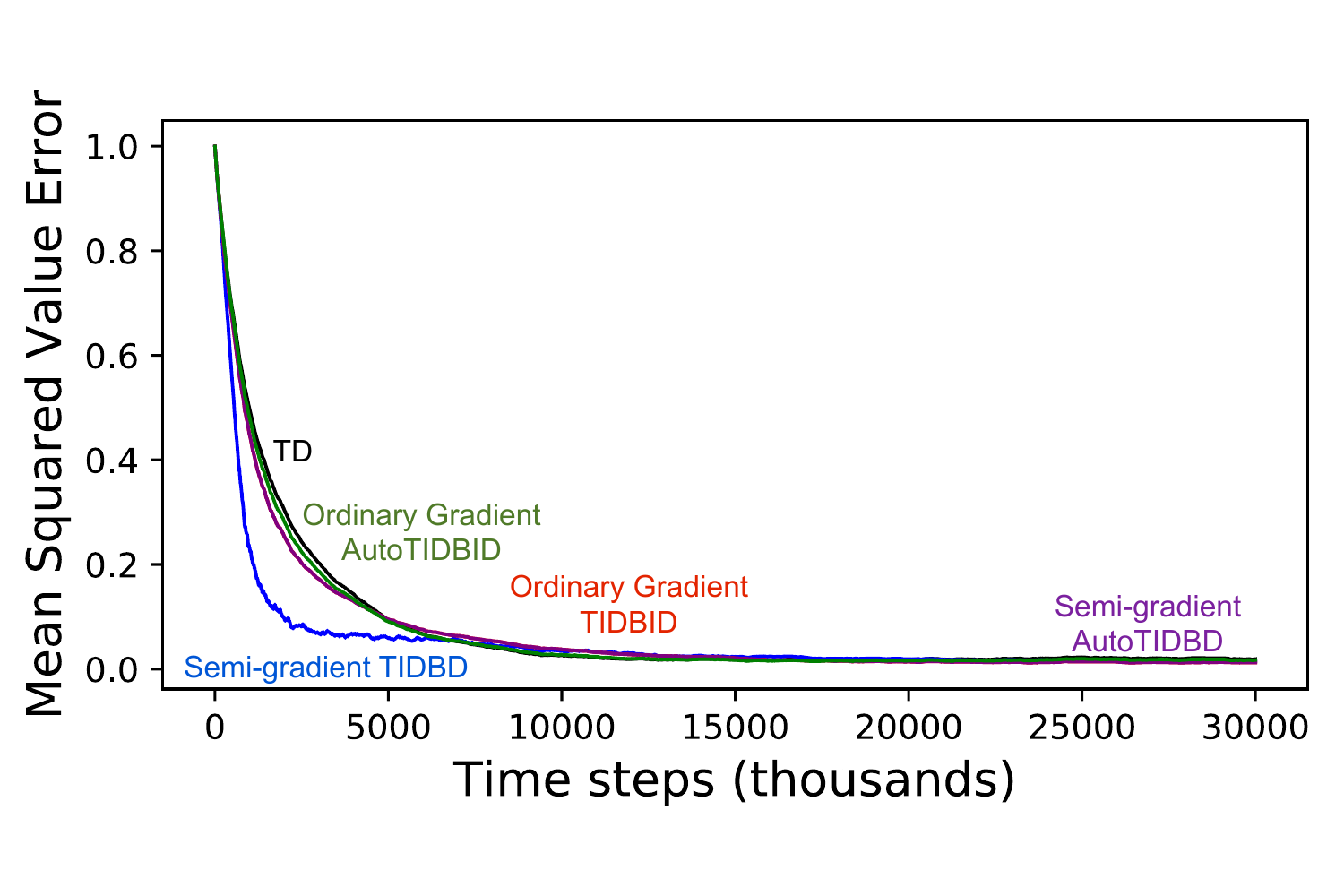}
\caption{$\alpha_0 = 0.05$}
\label{error_2_w_auto}
\end{subfigure}

\begin{subfigure}{\textwidth }
\centering
\includegraphics[height=2in, width=\linewidth, keepaspectratio]{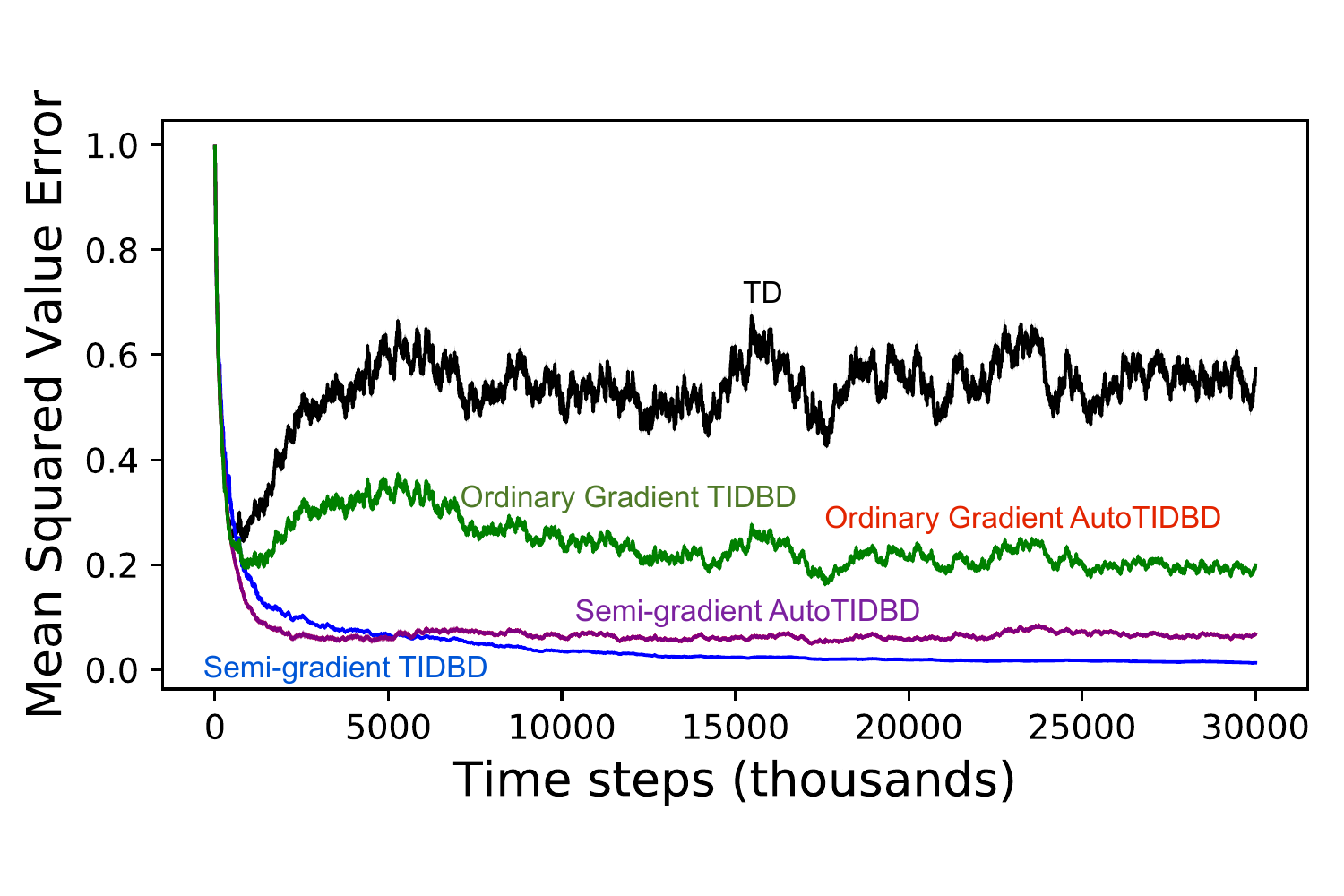}
\caption{$\alpha_0 = 0.5$}
\label{error_3_w_auto}
\end{subfigure}

\caption{Error during learning for three initializations of $\alpha_0$ for TD(0), both semi-gradient and ordinary-gradient TIDBD(0), and AutoTIDBD(0).}
\label{initial_learning_curves_auto}
% \end{minipage}%
\centering
\end{figure}

In Figure \ref{initial_learning_curves_auto}, the average error over the trial is shown for $\alpha_0$ of 0.01, 0.05, and 0.5: points corresponding to initializations of $\alpha_0$ as shown in Figures \ref{gridworld_autotidbd_full} and \ref{gridworld_autotidbd_semi}. We can see that for the more conservative initialization of $\alpha_0 = 0.01$, AutoTIDBD is able to increase its step size and learn faster than ordinary TD. As expected, AutoTIDBD performs about as well as ordinary gradient TIDBID without Auto normalization. When we start with a conservative step size, we are unlikely to diverge; AutoTIDBD and TIDBD will behave similarly in this setting. For the best initialization $\alpha_0 = 0.05$, AutoTIDBD performs as well as fixed step size TD. For the most aggressive setting $\alpha_0 = 0.5$, AutoTIDBD learns slower and more erratically than un-normalized TD.
% todo: explain erratic behaviour.

Unlike the previous versions of TIDBD,  AutoTDBD performs as well as or better than ordinary fixed step size TD for all initial settings of $\alpha$.  While this stability and improved performance comes at a cost---AutoTIDBD learns more slowly for aggressive $\alpha_0$ than TIDBD---even without representation learning AutoTIDBD is able to perform as well as or better than ordinary TD learning for all of the initial step sizes $\alpha_0$ in the grid-world experiment.

\section{How Robust is AutoTIDBD to Selection of Meta Step Size \texorpdfstring{$\theta$}?}
\label{experiment intro}
In previous sections, we demonstrated that TIDBD and AutoTIDBD tuning a single, shared step size was able to perform as well as or better than tuned ordinary TD. We now evaluate how well AutoTIDBD performs when using a vector of many step sizes---when it is performing representation learning. We evaluate this on a known, challenging,  real-world prediction of problem \citep{pilarski_dynamic_2012,pilarski_adaptive_2013, seijen_true_2014}.

\subsection{Robotic Prediction Task}
\label{prediction task}

\begin{figure}
    \centering
    \begin{subfigure}{0.45\textwidth}
        \centering
        \includegraphics[height=1.5in,width=\linewidth, keepaspectratio]{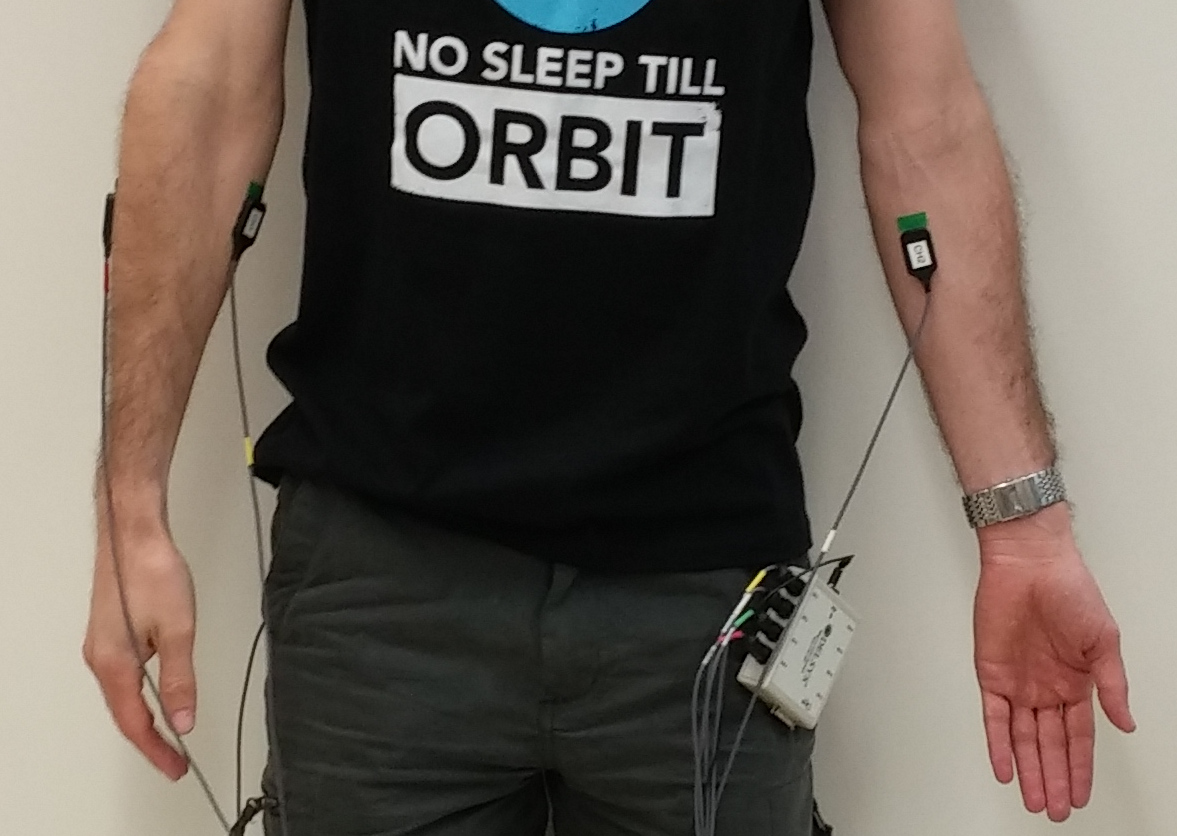}
        \caption{A subject with electrodes attached to their wrist flexors and extensors.}
        \label{electrodes}
    \end{subfigure}
    \begin{subfigure}{0.45\textwidth}
        \centering
        \includegraphics[height=1.5in,width=\linewidth,keepaspectratio]{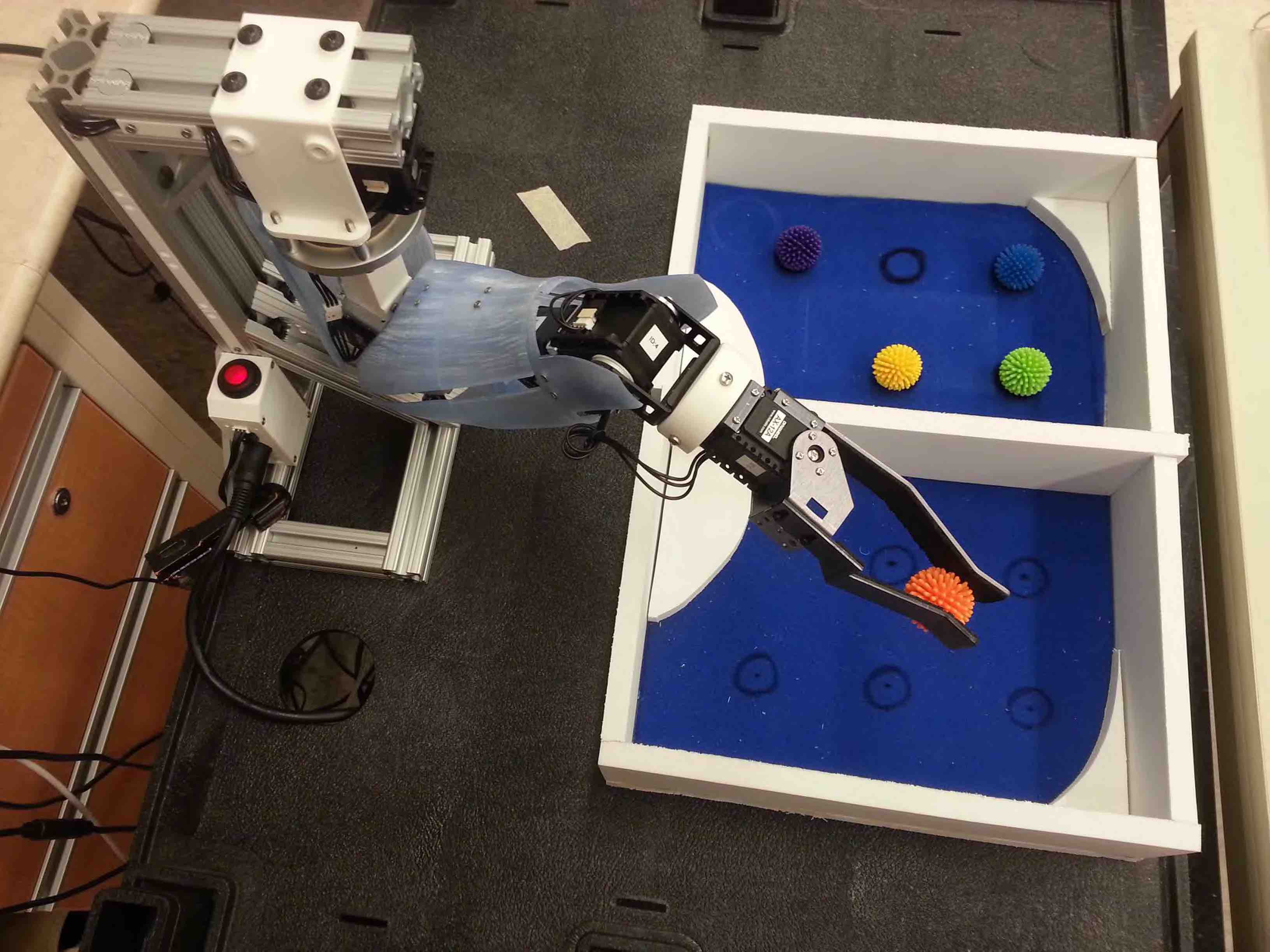}
        \caption{The BentoArm performing a modified Box and Blocks task.}
        \label{robot}
    \end{subfigure}
\end{figure}

\begin{figure*}[H]
\centering
\begin{subfigure}{0.45\textwidth}
    \centering
    \includegraphics[width=\linewidth]{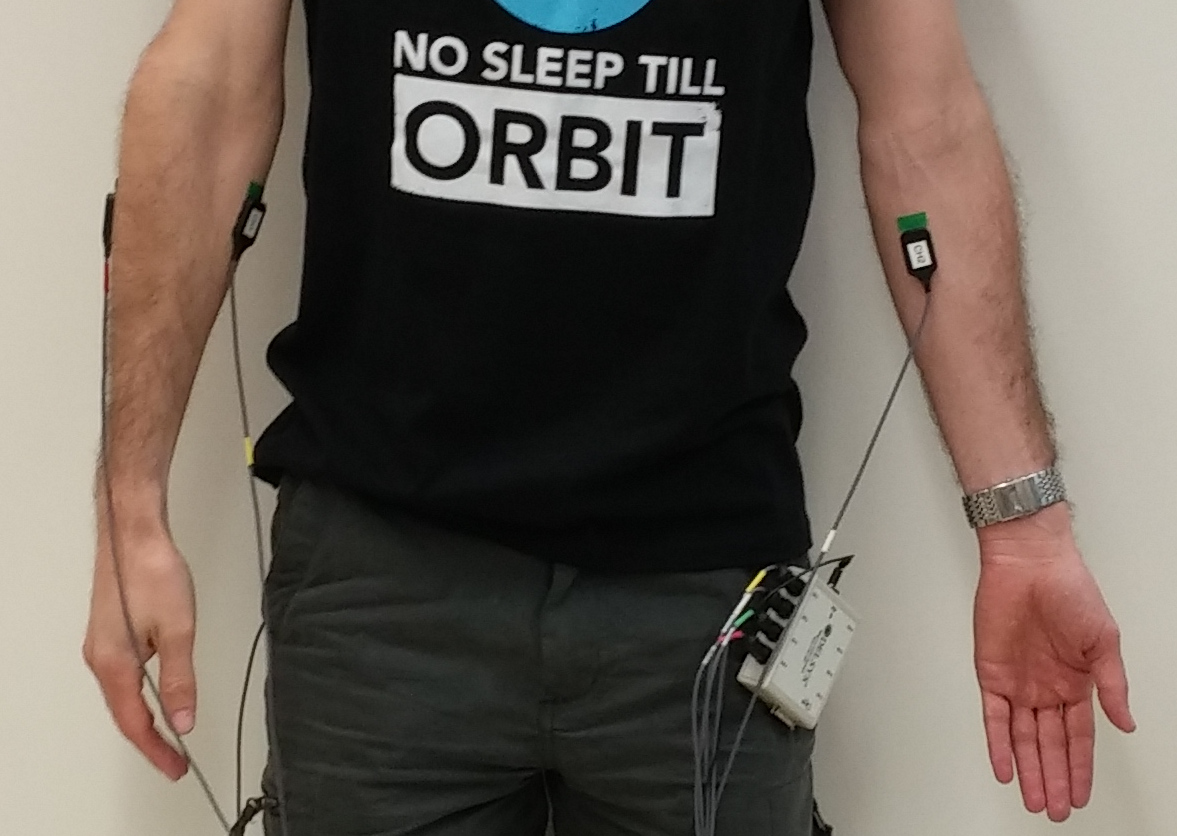}
    \label{electrodes}
    \caption{}
    
\end{subfigure}

\caption{Experiment setup for the robotic prediction task.}

\end{figure*}

The prediction problem for our evaluation consisted of predicting the temporally extended future values of signals of interest within the data stream of a robotic arm as a user controlled it to perform a manual manipulation task. The dataset for this evaluation was drawn from a prior study by \citet{edwards_machine_2016}, as done in the evaluation of True Online TD by \citet{seijen_true_2014}. In this dataset, the signal space of a robotic arm---the BentoArm of \citet{dawson_development_2014}, depicted in Figure \ref{robot}---was recorded as four participants used signals from their upper-arm muscles to control the robot to perform a simple object placement task. Signals in the data stream included the moment-by-moment position, velocity, load, temperature of all the robot's motors, along with the control signals being sent by the human user. Users were tasked with switching between the multiple controllable degrees of the robot arm to move balls from one side of a divided box to another. For the full experimental protocol used in generating this dataset, please refer to \citet{edwards_machine_2016}. Important to our present evaluations, in this dataset the four participants each performed the manipulation task a total of six times: three times with a non-adaptive control system, and three times with an adaptive control system, creating a total of 24 independent time-series trials within the dataset. The inclusion of data from the adaptive control case of \citet{edwards_machine_2016} adds non-stationarity to an already non-stationary prediction problem. Non-stationarity is introduced as the user becomes more proficient at the manipulation task, and as the control system begins to adapt to the user's preferred movements. No single, scalar step size would be ideal in this setting at all times, as the prediction problem changes over time. As noted by \citet{pilarski_adaptive_2013}, finding appropriate features and setting appropriate parameters for learning systems is in fact a known challenge in this particular robotic control domain---end-user time is precious, and designers cannot possibly test their prediction algorithms on datasets which are representative of all the situations the robot might encounter upon deployment. Aspects such as non-stationarity and the irregularities introduced through human-in-the-loop control therefore make this dataset an appropriate one for studying the robustness of AutoTIDBD in predicting a real-world data stream.

%For these reasons, finding appropriate features and choosing good parameter settings from experience is not only important for applications such as prosthetics, it is imperative.
%These electrodes recorded signals produced by muscles during contraction which were then used by a control interface to operate the robot. by flexing their arm in one direction, the subject could move the joint one way; by flexing their arm in the other direction, the joint would move in another direction. To change which joint they were operating on the limb, they would flex their other arm. This enabled the subjects to have bi-directional control of one joint on the limb the limb and an ability to switch between joints. For non-adaptive control interfaces, the order of list of joints which could be rotated through was fixed; for adaptive control interfaces, the order switched depending on which joint a system predicted a user would need next

% the timing between actions will change and the prediction of the hand position will shift.

% motivate the step size choice better

%These predictions are useful, as they can be used to inform more complex control systems which adapt to the user controlling the robot \citep{edwards_machine_2016,sherstan_collaborative_2015}, making it possible to continuously improve the control of these systems. 

\subsection{Sensitivity to Meta step size \texorpdfstring{$\theta$} for Prosthetic Prediction Problem}

For AutoTIDBD to be an improvement over ordinary, fixed step size TD, it should be less sensitive to settings of $\theta$ than TD is to settings of $\alpha$: less tuning should be required for AutoTIDBD. 

We constructed a prediction problem where each algorithm predicts a signal of interest from the robot arm. each algorithm predicted the angular position of the robot's hand motor (the gripper's aperture), as in \cite{seijen_true_2014}. We used tile-coding to construct a binary feature vector of size $2^{10}$ with 8 tilings and used the velocity of the hand, the position of the hand, and the participant's control signals to construct the feature vector. An additional bias feature was concatenated to the feature-vector, resulting in 9 active features at any given time. 

\begin{figure}[h]
\centering
\includegraphics[width=\linewidth, height=2.5in, keepaspectratio]{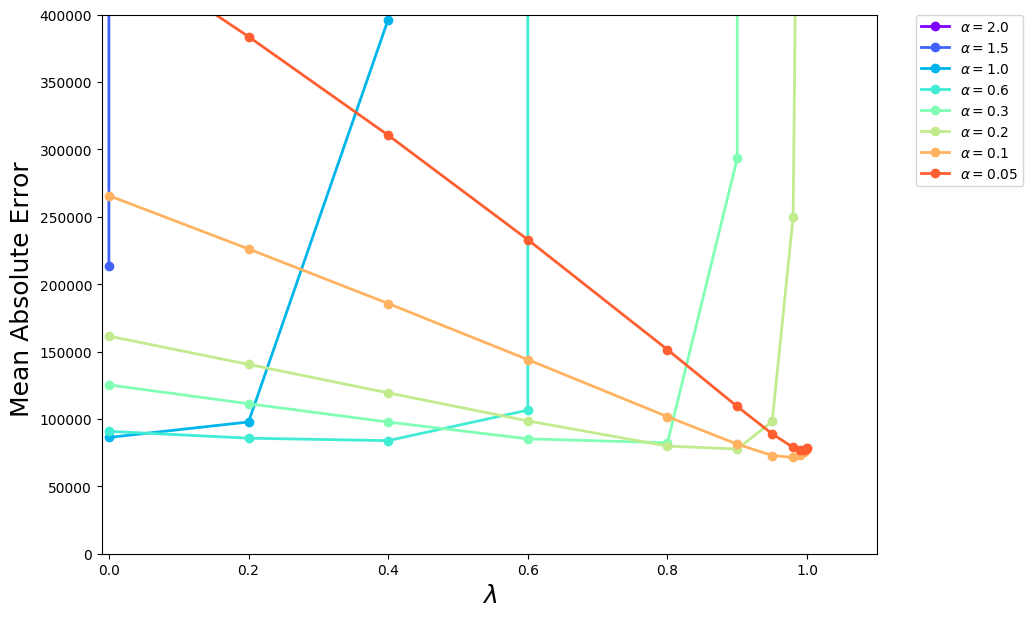}
\caption{Average cumulative error of TD($\lambda$) for various $\alpha$ settings.}
\label{step size-td}
\end{figure}

We compared AutoTIDBD to NoSID, SID, AutoSID, RMSProp, and AlphaBound, as described in source papers. Each learning method used a discounting factor $\gamma = 0.95$. The IDBD-based adaptive step size methods shown all initialized their vector step sizes to an initial value of $\log(\frac{1}{9})$, which when exponentiated results in a step size $ \alpha_0 = \frac{1}{9}$. This step size was chosen, as it was one of the best performing step size values for smaller values of $\lambda$, but diverges at $\lambda = 0.6$ for ordinary fixed shared step size TD($\lambda$) (shown in Figure \ref{step size-td}). AlphaBound is initialized with a step size of 1, as originally specified.

In figure \ref{step size-td}, the sensitivity of  static step size TD($\lambda$) is shown for a variety of $\alpha$ values across $\lambda$ settings. There is a trade-off between error and magnitude of step sizes. Consequently, there is no single step size $\alpha$ which performs well for all $\lambda$ values.

\begin{figure}[h]
\centering
\includegraphics[width=\linewidth, height=2.5in, keepaspectratio]{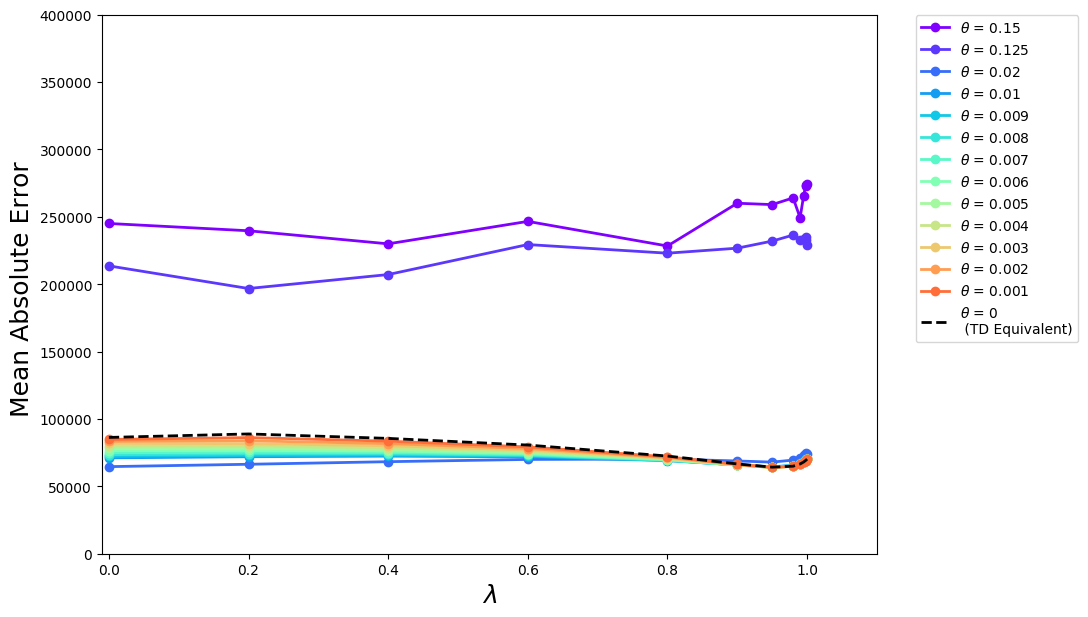}
\caption{Average cumulative return error of AutoTIDBD for various $\theta$ settings.}
\label{step size-auto}
\end{figure}

In Figure \ref{step size-auto} the sensitivity to the same prediction problem is shown. AutoTIDBD is less sensitive to to $\theta$ than TD($\lambda$) is to the setting of $\alpha$; AutoTIDBD performs well for each $\theta$ and $\lambda$ value. 
This satisfies our criteria for a method which performs acceptably for broad meta-parameter settings.

\begin{figure}[H]
\centering
\includegraphics[width=\linewidth]{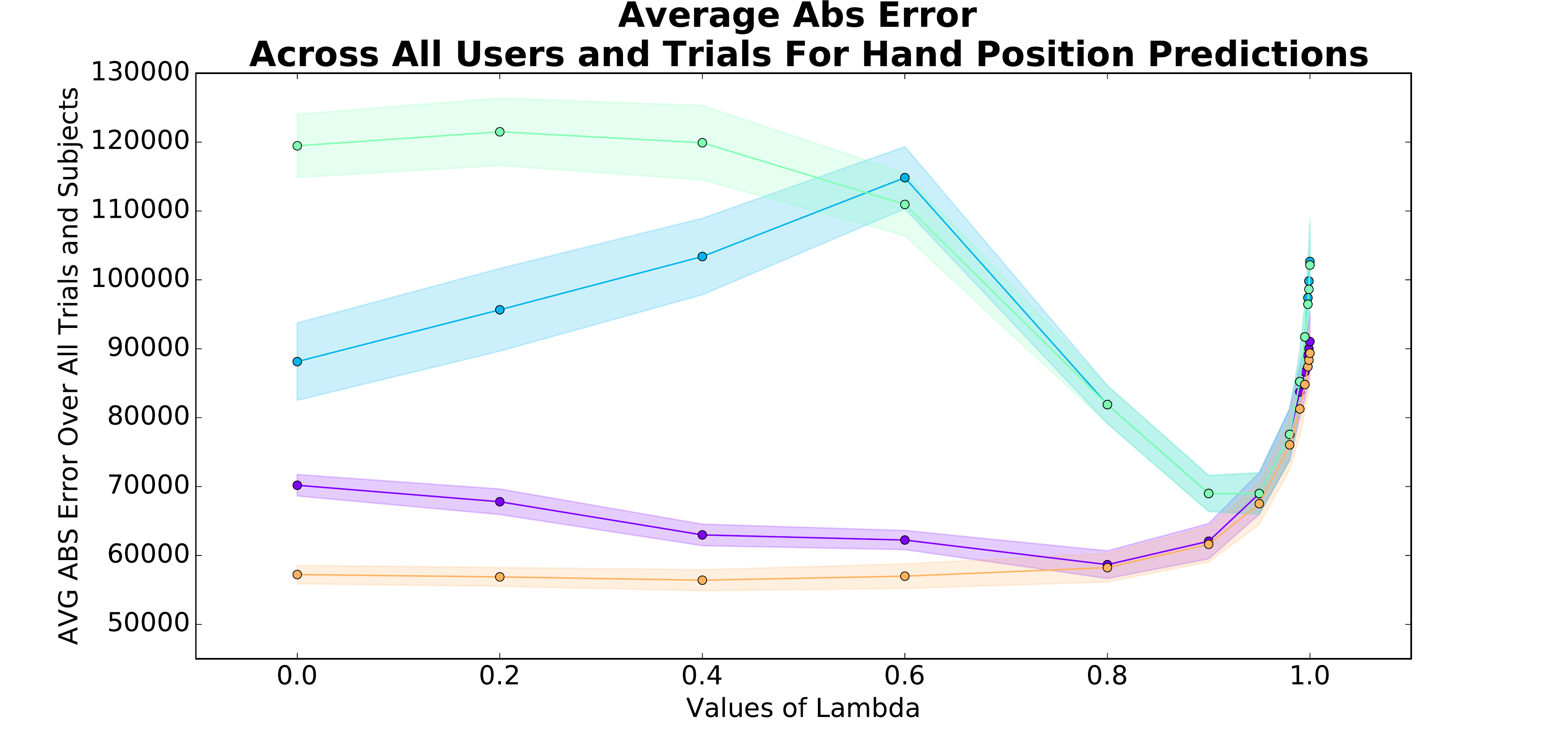}
\caption{Absolute cumulative return error averaged over 24 independent trials. Each algorithm is presented for the best setting of $\alpha$ or $\theta$ and is compared by varying $\lambda$ between 0 and 0.9}
\label{comparison}
\end{figure}

In Figure \ref{comparison}, the average cumulative error for the best tuned parameter settings for each method are shown. Adaptive step size we compared against are SID and Alphabound. These methods act as a baseline comparison. If TIDBD and AutoTIDBD are performing representation learning well, we expect them to perform better than scalar step size adaptation methods. TD with RMSProp, AutoSID, and NOSID could not be compared, as both had errors which were too large to be compared.

Both Semi-gradient TIDBD and Ordinary gradient TIDBD outperform SID---A scalar version of IDBD for TD learning; however, neither consistently attain errors less than or equal to ordinary TD. In contrast, AutoTIDBD  attains lower or equivalent error to TD---outperforming all other adaptive methods we compared against, excluding AlphaBound for large values of $\lambda$. 

\subsection{Sensitivity to Meta step size \texorpdfstring{$\theta$} Across Prediction Problems}

We previously assessed the performance of AutoTIDBD across meta step size settings for predictions of the gripper. An ideal adaptive step size method should have broad ranges of meta-parameter settings for which it attains acceptable performance. Moreover, it is ideal if these meta-parameter settings are invariant over different problems: if an ideal step size on one problem is ideal for all other problems.

Using the experiment and setup from Section \ref{experiment intro}, we compare the sensitivity of AutoTIDBD and TIDBD with an ordinary gradient for a variety of prediction formulations. We predict the velocity, position, and load for all five servos of the robot arm. Each of the signals produced by the arm are unique; as a result, each of the prediction problems is unique.

% todo: make a graph visualizing how different all the servos from this arm are. 

\begin{figure}[H]
\includegraphics[width=\linewidth]{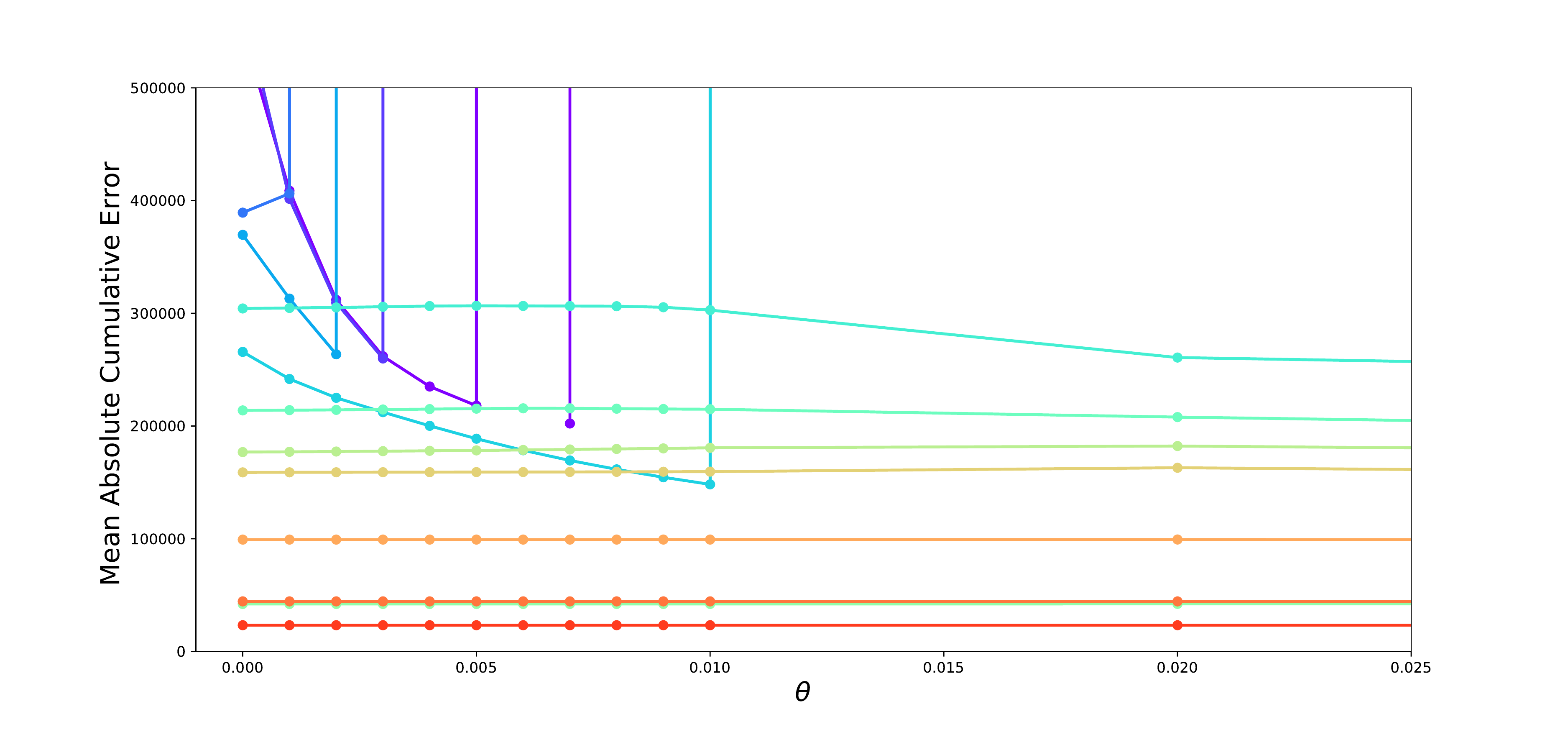}
\caption{Average absolute cumulative return error of TIDBD(0) for different values of $\theta$. Each line represents the error of a prediction with its own unique signal of interest. Each on-policy prediction has the same $\gamma$, but is predicting a different signal of interest.}
\label{meta_tidbd}
\end{figure}

In Figure \ref{meta_tidbd}, the sensitivity of ordinary gradient TIDBD to it's meta-parameter settings is depicted. Each line represents a prediction of a different signal of interest from the arm. We can see that the sensitivity for each problem is different and the best setting of $\theta$ for each problem are different. This is problematic---the problem of tuning step sizes has simply been abstracted away to a higher-level. TIDBD may be less sensitive to initialization of  its parameters than ordinary TD, but it is still sensitive and thus must be tuned for each target domain.

\begin{figure}[H]
\includegraphics[width=\linewidth]{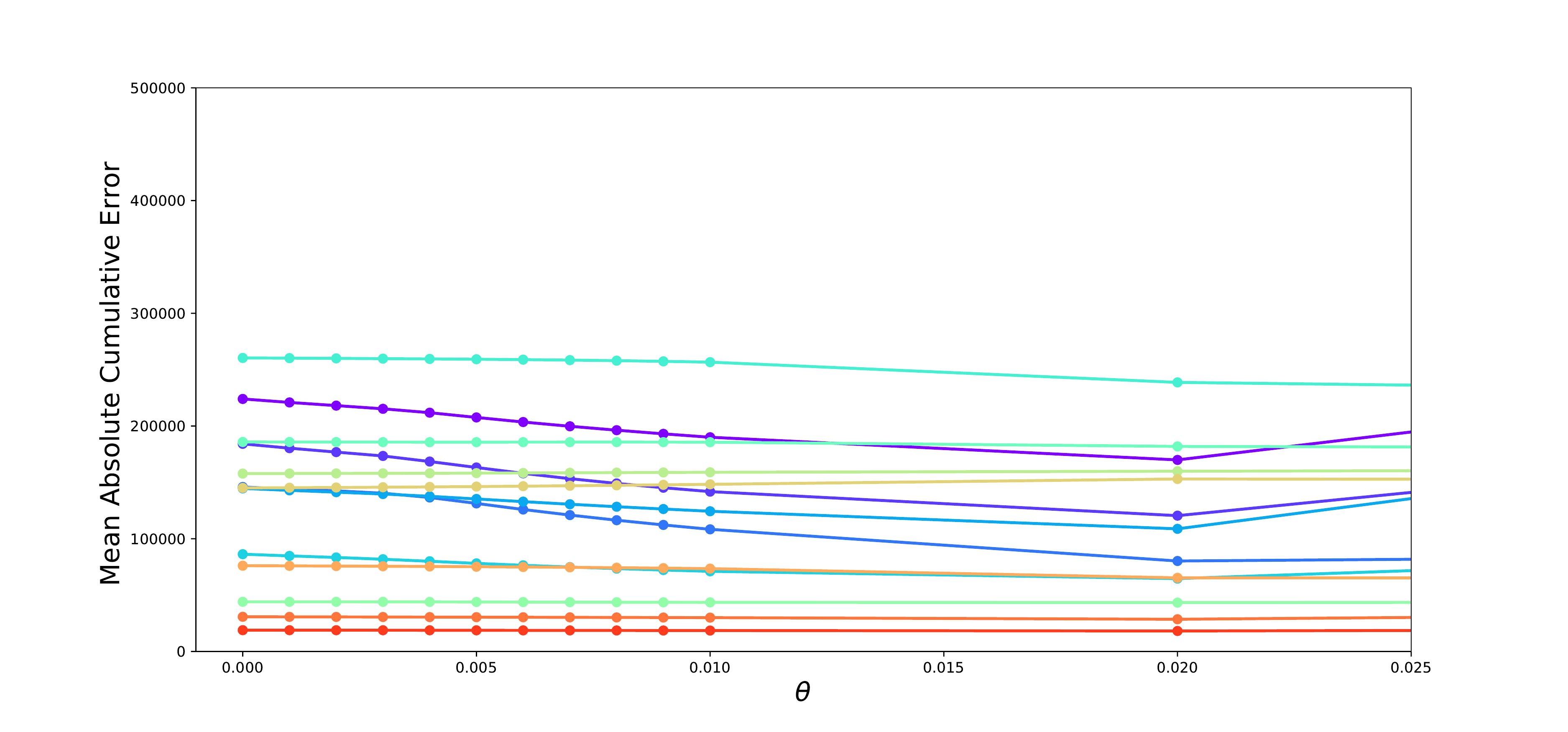}
\caption{Average absolute cumulative return error of TIDBD(0) for different values of $\theta$. Each line represents the error of a prediction with its own unique signal of interest. Each on-policy prediction has the same $\gamma$, but is predicting a different signal of interest.}
\label{meta_auto_tidbd}
\end{figure}

In Figure \ref{meta_auto_tidbd} the same meta-sensitivity for AutoTIDBD is depicted for the same meta-parameter settings. Each of the settings eventually diverge, but the valley of $\theta$ parameters for which we attain reasonable performance is far broader than TIDBD and TD($\lambda$), and relatively invariant across prediction problems.

AutoTDBID meets our requirements: It performs as well as or better than ordinary TD($\lambda$)---even when only adapting a single step size, it is less sensitive to settings of $\theta$ than TD($\lambda$) is to $\alpha$, and has broad ranges of $\theta$ which are invariant across problems. AutoTIDBD is bias-adaptation step size method which does not require tuning across applications. What remains of our criteria to be evaluated is whether AutoTIDBD is capable of performing representation learning.

\section{Can AutoTIDB Perform Representation Learning?}

We have demonstrated that AutoTIDBD is able to outperform scalar step size adaptation methods and ordinary TD on real-world prediction problems by tuning its step sizes, and that it is less sensitive to its parameters than ordinary TD. What remains of our four criteria is to determine if TIDBD is able to effectively perform representation learning by giving features relevant to the current prediction task large step sizes, and small step sizes to features which are irrelevant. We assess AutoTIDBD's ability to perform representation learning, by analyzing the change in step size values for the prediction task introduced in Section \ref{experiment intro} at $\lambda = 0.95$. 

% {\bf TODO: more detail needed on how random features were created and updated, so that an independent reader could replicate the experiment.} 

We created poor features by randomly choosing 25\% of the features to be noisy. Noisy features were activated equiprobably. After completion of the experiment, the noisy features were compared with the number activations of each feature to ensure that noisy features included some which were highly active. There are certain time-steps for which all of the step sizes suddenly decrease. On these time-steps, the effective step size was greater than 1, leading to the normalization of the meta-weights to prevent over-shooting.

\begin{figure}[H]
\centering
\includegraphics[width=\linewidth, height =2.7in,keepaspectratio]{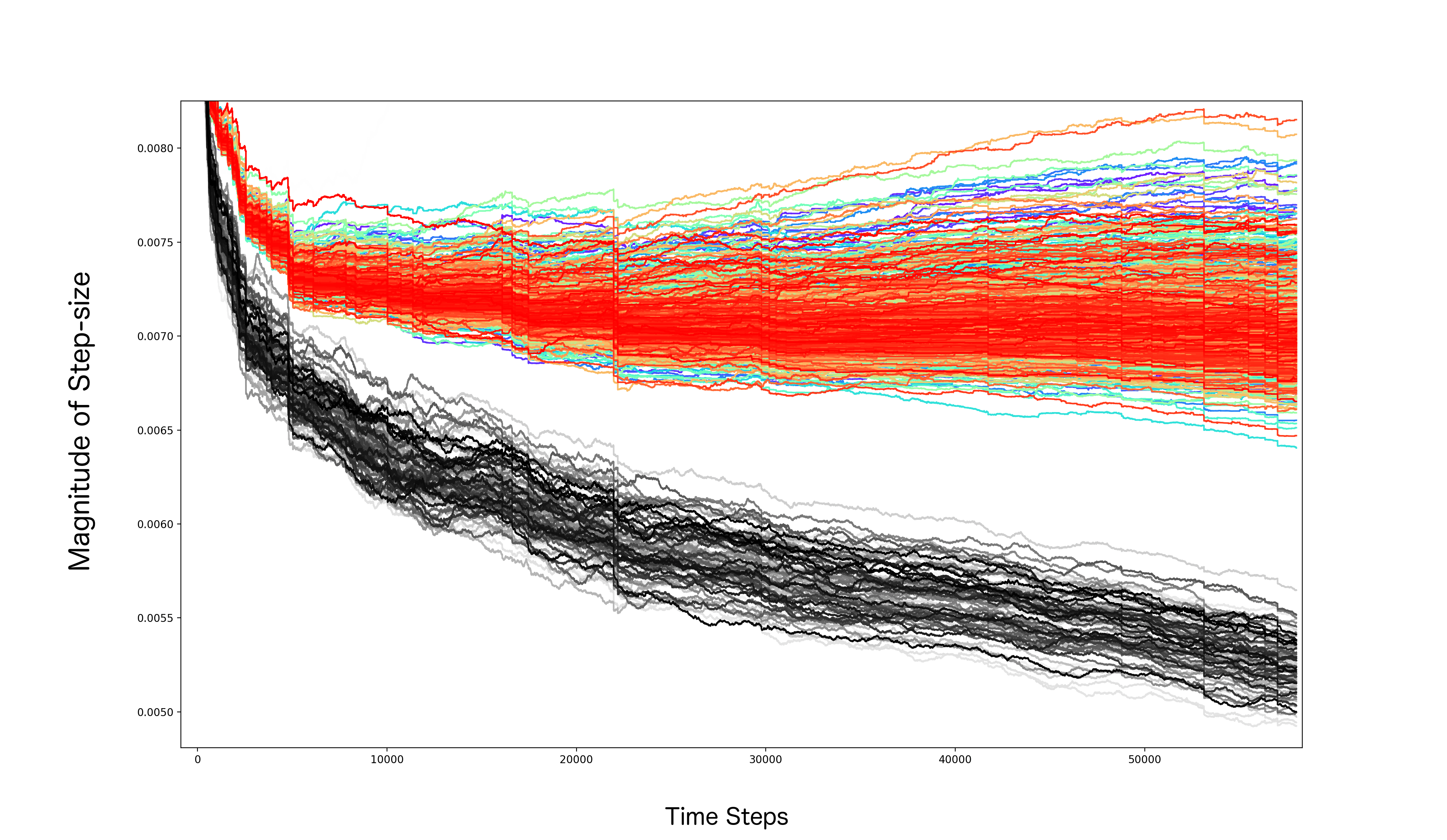}
\caption{Average magnitude of step sizes over all trials. Noisy features are in greyscale, ordinary features are in colour. }
\label{noise}
\end{figure}

Figure \ref{noise} depicts the magnitude of all step sizes averaged over all trials. step sizes corresponding to noisy features consistently decrease over time. We can see that the noisy features coloured in greyscale constantly shrink as experience determines them to be unreliable. This creates a separation between features which are noisy and those which are not. None of the noisy features have values within the range of ordinary features---all of the noisy features were correctly given smaller step sizes.

AutoTIDBD is able to perform representation learning by assigning appropriate step sizes, meeting our final criteria. 

\section{Limitations and Future Work}

This derivation of TIDBD is limited to methods which use linear function approximation. Generalizations of IDBD for non-linear supervised learning methods exist \citep{schraudolph_local_1999}, and could be generalized to TD learning in the future. In addition, this generalization is limited to on-policy predictions with replacing and accumulating traces. Further extension is required before AutoTDBD can be used with off-policy prediction methods and, and methods with different eligibility traces such as True Online TD \citep{seijen_true_2014}.

In addition to producing more general methods, future research could pursue additional uses of learned step sizes. Step sizes learned with IDBD methods describe the relevance of given feature to the task at hand. If many of the features are large, then the features for the given prediction are well specified. This possibly be used to evaluate the potential of a prediction based on its given feature representation before it has been completely learned. Preliminary evaluation of predictions based on step sizes could be beneficial to prediction architectures such as horde \citep{sutton_horde_2011}, where large collections of predictions proposed, learned, and maintained in real-time as a learner is interacting with their environment. In such situations limited computational resources must be used effectively; being able to better identify promising predictions in early learning could support prediction discovery in predictive knowledge systems.

Learned step sizes may also be an effective way to drive computational curiosity and intrinsic motivation. A challenge for learning systems is deciding how to explore their environment to support learning. Many intrinsic motivation systems rely on metrics which drive exploration based on error on a given task \citep{oudeyer_what_2009}. One shortfall of these approaches is difficult for these methods to differentiate between situations where the error is high because not enough learning has occurred, and situations where the error is high because some signal or portion of the environment is not learnable. Learned step sizes describe how much learning, if used in combination with traditional error-based forms of intrinsic motivation, it may be better able to differentiate between what is novel and should be learned about , and what is unlearnable.

\pagebreak

\section{Conclusion}

We presented an approach to generalizing Incremental Delta-Bar-Delta to temporal-difference learning, demonstrating that its effectiveness carries over from supervised learning to TD. We extended TIDBD to AutoTIDBD, using normalization methods from Autostep to improve the robustness of TIDBD. Adapting step sizes with AutoTIDBD is an improvement over ordinary TD methods with a tuned static step size, even on stationary problems. On non-stationary tasks, we showed that AutoTIDBD is able to find appropriate step sizes and differentiate between relevant and irrelevant featdures. Most importantly, over a number of real-world robotic prediction tasks we demonstrated that AutoTIDBD is less sensitive to choices of meta step sizes $\theta$ and initial step sizes $\alpha_0$ than ordinary TD is to settings of $\alpha$. AutoTIDBD out-performs TD for broad a broad range of meta step size settings which is relatively invariant over prediction problems. AutoTIDBD and TIDBD-based step size learning systems show promise of learning feature relevance and performing meta learning in an incrementally and online, lessening dependence on feature construction and parameter tuning.

\section*{References}
%\pagebreak
\bibliographystyle{elsarticle-harv} 
\bibliography{My_Library}

% \end{thebibliography}
\end{document}